\pdfoutput=1

\documentclass[11pt]{article}

\usepackage[final]{acl}

\usepackage{times}
\usepackage{latexsym}

\usepackage[T1]{fontenc}

\usepackage[utf8]{inputenc}

\usepackage{microtype}

\usepackage{inconsolata}

\usepackage{multirow}
\usepackage{booktabs}
\usepackage{graphicx}
\usepackage{afterpage}
\usepackage{bbding}
\usepackage{lineno}
\usepackage{amsmath}
\usepackage{amssymb}
\usepackage{enumitem}
\usepackage{tikz}
\usepackage{bm}
\usepackage{seqsplit}
\usepackage{xcolor}
\usepackage{array}

\definecolor{mygreen}{HTML}{00B050}
\definecolor{myred}{HTML}{FF0000}

\title{ESCoT: Towards Interpretable Emotional Support Dialogue Systems}

\author{\textbf{Tenggan Zhang${^{1,\ast}}$, Xinjie Zhang${^{1,\ast}}$, Jinming Zhao$^2$, Li Zhou$^3$, Qin Jin${^{1,\dagger}}$} \\
  \small $^1$School of Information, Renmin University of China \\
  \small $^2$Independent Researcher \\
  \small $^3$Mental Health Education and Counseling Center, Renmin University of China \\
  {\small \tt zhangtenggan@gmail.com, zhangxinjie827@ruc.edu.cn, }\\
  {\small \tt zhaojinming1@gmail.com, psyzhouli@ruc.edu.cn, qjin@ruc.edu.cn} \\
}

\begin{document}
\maketitle

\def\thefootnote{$\ast$}\footnotetext{Co-first authors with equal contribution.}
\def\thefootnote{$\dagger$}\footnotetext{Corresponding Author.}

\begin{abstract}
Understanding the reason for emotional support response is crucial for establishing connections between users and emotional support dialogue systems. 
Previous works mostly focus on generating better responses but ignore interpretability, which is extremely important for constructing reliable dialogue systems. 
To empower the system with better interpretability, we propose an emotional support response generation scheme, named \textbf{E}motion-Focused and \textbf{S}trategy-Driven \textbf{C}hain-\textbf{o}f-\textbf{T}hought (\textbf{ESCoT}), mimicking the process of \textit{identifying}, \textit{understanding}, and \textit{regulating} emotions. 
Specially, we construct a new dataset with ESCoT in two steps:
(1) \textit{Dialogue Generation} where we first generate diverse conversation situations, then enhance dialogue generation using richer emotional support strategies based on these situations;
(2) \textit{Chain Supplement} where we focus on supplementing selected dialogues with elements such as emotion, stimulus, appraisal, and strategy reason, forming the manually verified chains. 
Additionally, we further develop a model to generate dialogue responses with better interpretability.
We also conduct extensive experiments and human evaluations to validate the effectiveness of the proposed ESCoT and generated dialogue responses.
Our data and code are available at \href{https://github.com/TeigenZhang/ESCoT}{https://github.com/TeigenZhang/ESCoT}.
\end{abstract}

\section{Introduction}

\begin{figure}[t]
\centering
\includegraphics[scale=0.8]{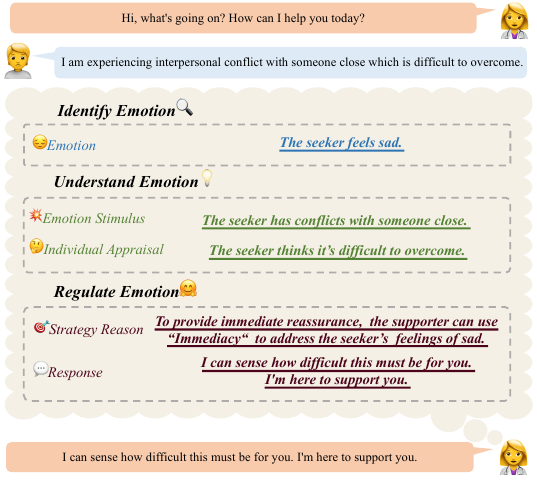}
\caption{Illustration of the ESCoT scheme.
The supporter first \textit{identifies emotion}, then \textit{understands emotion} from perspectives of emotional stimulus and individual appraisal, and finally chooses the appropriate strategy and responds to the seeker to \textit{regulate emotion}.} 
\label{fig:escot}
\end{figure} 

Emotional support is conceptualized as expressing care, concern, affection, and interests, especially for the individuals feeling stressed or upset~\cite{burleson2003emotional,cdata1987communicating,cutrona1987provisions}. 
Incorporating emotional support can yield positive effects in many scenarios, such as therapeutic sessions~\cite{kennelly2001music}, customer service counters~\cite{barnes2005enhancing}, and palliative cares~\cite{skilbeck2003emotional}.
Realizing reliable emotional support dialogue systems capable of automating these interactions is expected to expand the scope and efficacy of such services.
Moreover, a reliable emotional support dialogue system should not work like a black box, providing conversational responses but unable to explain how those responses were generated~\cite{xAI}. 
As shown in Figure~\ref{fig:escot}, let's imagine how a helpful supporter would work by considering the feelings of a seeker who asks for help: the supporter would first \textit{identify} the situation and the emotion of the seeker, then \textit{understand} and acknowledge the emotion, and finally choose appropriate strategies to respond in order to \textit{regulate} the emotion~\cite{j1996peer}.
Therefore, it is extremely desired to build a reliable and trustworthy emotional support dialogue system that can not only generate emotional support responses but also provide the reasoning or chain-of-thought (CoT) behind the generated responses. 

Some previous endeavors have attempted to improve the interpretability of emotional support dialogue systems, such as controlling the response by emotion~\cite{FuZWM23,emotion_cause_empathy} or strategy~\cite{ChengLLWZLL022,WelivitaP23}, or using commonsense to augment the emotional support response~\cite{EEDG-PCI,CaiSXSWGZX23}. However, to the best of our knowledge, there is currently no such emotional support dialogue system that can provide comprehensive reasoning explanations. Therefore, in this work, we aim to build an interpretable emotional support dialogue system. 

Due to the high expertise requirements for supporter roles in emotional support conversations, building a human-annotated emotional support dialogue dataset is very costly. Recently, the powerful language generation and reasoning capabilities of large language models (LLMs) have demonstrated a viable pathway to generate high-quality data. Efforts such as \textsc{AugESC} \cite{augesc} and \textsc{SmileChat} \cite{smileconv} have attempted to expand emotional support dialogue datasets via LLMs. However, the reasoning or chain-of-thought behind the dialogue responses has been overlooked.

In this paper, we propose
an emotional support response generation scheme, named \textbf{E}motion-focused and \textbf{S}trategy-driven \textbf{C}hain-\textbf{o}f-\textbf{T}hought (\textbf{ESCoT}), to generate dialogue data, inspired by the human emotional support generation process of \textit{identifying}, \textit{understanding}, and \textit{regulating} emotions.
Specifically, to emphasize the critical role of conversation strategies and dialogue situations, we first create diverse dialogue situations, and then enhance dialogue generation using richer emotional support strategies based on these situations. Furthermore, we complement selected dialogues with the chain-of-thought (CoT), which is represented as a quintuple \((EM, ES, IA, SR, RE)\), reflecting the process as illustrated in Figure~\ref{fig:escot}. 
After careful manual checking, we build the first dataset for \textbf{E}motional \textbf{S}upport \textbf{D}ialogue with \textbf{CoT} (\textbf{ESD-CoT}), containing 1.7k+ dialogues. 
Moreover, we build our emotional support dialogue system with better interpretability via supervised fine-tuning a pre-trained language model on ESD-CoT, providing a strong baseline for future investigation.

Our main contributions in this work include: (1) We develop an effective emotion-focused strategy-driven chain-of-thought automatic data generation scheme called ESCoT to increase the interpretability of emotional support response generation. 
(2) We build the first chain-of-thought emotional support dataset ESD-CoT, containing 1.7k+ dialogues through automatic generation and manual correction.
(3) We conduct human evaluations to validate from different aspects the effectiveness of our data generation scheme and the quality of our constructed dialogue dataset. 
(4) We build an interpretable emotional support response generation model on ESD-CoT and conduct a comprehensive assessment of the performance, providing a strong baseline for future research.

\section{Related Work}

\paragraph{Datasets Associated with Emotional Support}
Lack of sufficient datasets is one of the challenges faced by emotional support dialogue systems.
Due to strict personal privacy protection requirements and high expertise demands, constructing high quality and diverse empathetic and emotional support dialogue datasets is extremely challenging for humans. 
\citet{SharmaMAA20} construct a dataset for supporting the \textsc{Epitome} model based on TalkLife and Mental Health Subreddits.
\citet{psyqa} scrape Q\&A from the Yixinli platform and annotate responses based on psychological counseling theories to create the PsyQA Chinese dataset.

These datasets are non-dialogue datasets, but applications like psychological counseling need multi-turn dialogues.
\citet{empatheticdialogues} propose a new emotional dialogue generation benchmark and create a new dataset called \textsc{EmpatheticDialogues}, which contains 25k dialogues in emotional contexts. 
\citet{esconv} propose the ESC framework and construct a dataset named ESConv based on this theoretical framework.

In order to obtain such datasets at a lower cost, some works leverage the power of LLMs to augment data for emotional support dialogues. 
\citet{augesc} introduce \textsc{AugESC}, an augmented dataset for the ESC~\cite{esconv} task by leveraging fine-tuned large language models to complete dialogues.
\citet{smileconv} develop the SMILE approach by using ChatGPT~\cite{ChatGPT2022} to transform single-turn dialogues into multi-turn conversations.

However, these works merely treat data augmentation as dialogue continuation or rewriting tasks, without making specific adjustments to accommodate the features of emotional support dialogues. For instance, dialogue situations and conversation strategies have not been taken into consideration.

\paragraph{Interpretable Dialogue Systems}

Deep learning models for dialogue systems are often seen as black boxes due to the complexity and opacity of their internal mechanisms. Lack of interpretability can lead to safety concerns, as it's challenging to predict or understand the models' decisions in critical scenarios. To address these concerns, researchers have been exploring different methods to improve the interpretability of language models.

One research direction involves integrating knowledge-based reasoning to improve the moral and ethical judgment capabilities of dialogue systems. ~\citet{MehrabiBMG22} and ~\citet{0002YJLKKCS22} try to incorporate external knowledge sources and structured reasoning pathways to enhance the decision-making quality of these models, particularly in scenarios requiring moral or ethical considerations. 
~\citet{LiSXTLGSJW23} introduce an interpretable dialogue system that employs a two-stage response generation process, enhancing response diversity and system transparency.
Moreover,~\citet{cbtllm} create dialogue modules based on CBT~\cite{beck1979cognitive} dialogue scenarios centered on Socratic questioning and consider the questions about ABC~\cite{ellis1991revised}. 
An emerging direction is to explain language models using LLMs.
~\citet{bills2023language} explore using GPT-4~\cite{gpt4} to interpret and understand the behavior of neurons in language models such as GPT-2 XL~\cite{gpt2}.
However, there is currently no emotional support dialogue system that can provide comprehensive reasoning explanations to improve the interpretability.

\paragraph{Chain-of-Thought Prompting}
\citet{wei2022chain} initially introduce Chain-of-Thought (CoT) Prompting to mimic the reasoning process. 
Following this, various works that utilize CoT to prompt LLM for intricate reasoning tasks spring up across different domains, such as Auto-CoT~\cite{zhang2022automatic}, SP-CoT~\cite{wang2023self}, and PsyCoT~\cite{yang2023psycot}. 
To address challenges when applying CoT prompting in dialogues, Dialogue CoT~\cite{chae2023dialogue} propose decomposing commonsense reasoning into steps and generating rationale as a sequence of inferred commonsense knowledge required for response generation to facilitate Dialogue CoT reasoning. 
The Cue-CoT~\cite{wang2023cue} prompts the system to infer the user status first and then generate a response based on dialogue context and user status. 
However, CoT for emotional support dialogue systems has not been well explored yet.

\section{ESD-CoT Dataset Construction}

\begin{figure}[t]
\centering
\includegraphics[scale=0.48]{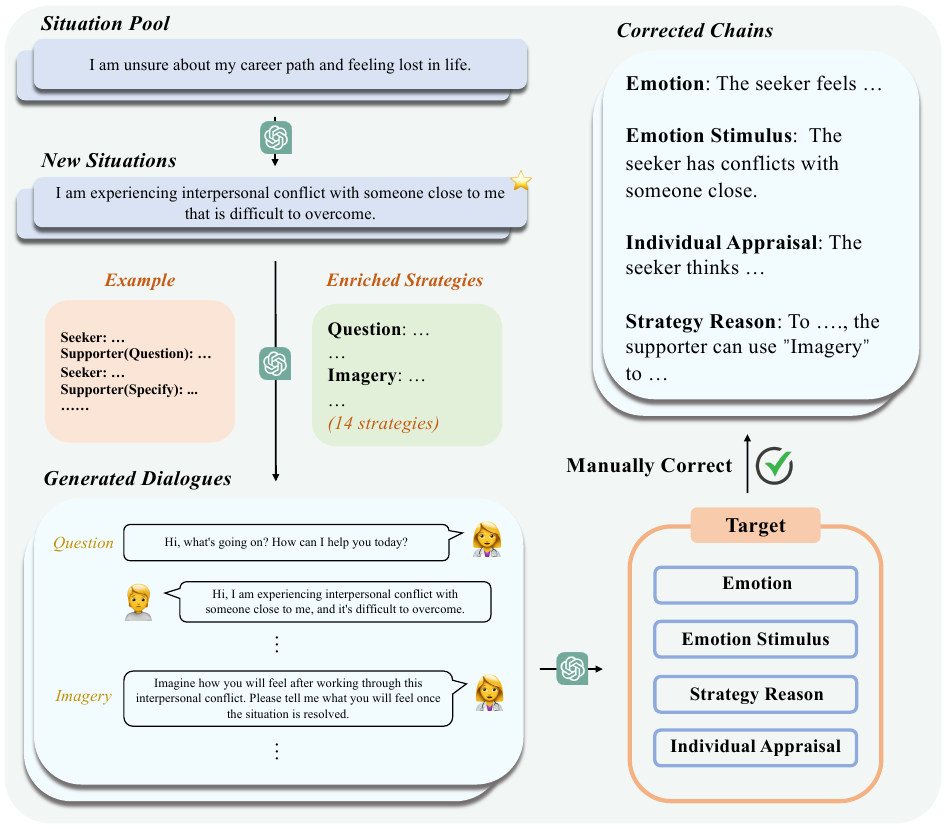}
\vspace{-8pt}
\caption{Illustration of our data generation scheme. We construct the ESD dataset according to the left-side process, and subsequently build the ESD-CoT dataset following the quintuple of \( (EM, ES, IA, SR, RE)\) in the right-side process.
}
\label{fig:pipeline}
\end{figure}

Due to the high expertise requirement in the supporter role, manually constructing an emotional support dialogue dataset is very costly.
Mimicking the human thought process of comforting others, we propose an Emotion-focused and Strategy-driven Chain-of-Thought (ESCoT) scheme to generate emotional support responses in an interpretable manner, and build the first Emotional Support Dialogue with CoT dataset (ESD-CoT).

The construction process of ESD-CoT dataset can be divided into two steps: 
(1) \textit{ESD Construction} where we first create diverse situations, and then generate dialogues with enriched strategies based on generated situations by leveraging LLMs; 
(2) \textit{ESD-CoT Construction} where we first generate reasoning chains of the selected dialogues via LLMs, and then conduct manual verification and modification to ensure the accuracy of the chains. 

\subsection{ESD Construction}
Considering the critical role of situation and strategy in emotional support dialogues, we propose a situation- and strategy-guided dialogue generation scheme based on large language models to build a situation-diverse, strategy-rich Emotional Support Dialogue dataset (ESD). Specifically, we first generate a more diverse range of psychological counseling situations and enrich the existing strategies used in \cite{esconv}. Then, we generate emotional support dialogues with the guidance of different situations and enriched emotional support strategies. 
Furthermore, we conduct extensive data analysis to demonstrate the quality of the generated data, and also conduct human evaluation to validate the necessity of strategy enrichment.

\subsubsection{Situation Generation}
The situations reflect the issues for which seekers are asking for help. In order to produce high-quality emotional support dialogues, various realistic psychological counseling situations are crucial. Inspired by \cite{self-instruct}, we leverage the in-context learning ability of ChatGPT for extensive and diverse situation generation.

We initialize the original situation pool with 1,300 manually annotated situations in ESConv~\cite{esconv}, forming a seed pool. We use the in-context learning method and design a situation generation prompt for ChatGPT.
For each generation iteration, we randomly select eight situations from the seed pool as in-context examples and generate eight new situations adding to the seed pool. More details of the situation generation prompt are presented in Appendix~\ref{Appdix:situation-prompt}.

To ensure high-quality and diversity of the generated situations, 
we remove duplicate situations and filter out inadequate situations that lack personal pronouns or have incomplete sentences etc. 
Finally, 2,943 new situations are retained to enrich the generation of subsequent dialogue data.

\subsubsection{Strategy Enrichment}
Eight strategies are employed in the ESConv dataset~\cite{esconv}, while other important strategies which are useful for emotional support have not yet been employed.
Considering the significance of strategies in practical counseling, we therefore are motivated to further enrich the existing eight strategies established by ESConv. 

Based on suggestions from experienced psychological counselors, we enrich the strategies following three principles:
(1) \textbf{\textit{Distinct}}: make sure each strategy focuses on different aspects compared to existing strategies.
(2) \textbf{\textit{Understandable}}: make sure each strategy is concise and comprehensible, even if it does not come with a short description.
(3) \textbf{\textit{Identifiable}}: make sure the implementation of the strategy can be easily identified from a few sentences.
Following the above principles and under the guidance of experts, we extract six strategies from helping skills~\cite{hill2009helping}, including \textit{Summarize}, \textit{Imagery}, \textit{Specify}, \textit{Take Responsibility}, \textit{Homework Assignment}, and \textit{Immediacy}. 

Note the focus of different strategies varies. For example, \textit{Summarize} is a general summary of the whole conversation, while \textit{Restatement or Paraphrasing} focuses on a simple restatement of the content just mentioned. \textit{Homework Assignment} is a type of direct guidance, which directly tells the seeker what to do, while \textit{Providing Suggestions} doesn't directly tell the seeker what to do.
Through the expanded diverse strategies, we can generate higher quality dialogues, demonstrated by the human evaluation in \ref{strategy_impact_assessment}. 
Detailed definitions and examples of these enriched strategies are provided in the Appendix~\ref{Appdix:strategy}.

\subsubsection{Dialogue Generation}
After obtaining sufficient situations and richer emotional support strategies, we proceed to generate emotional support dialogues based on these situations and strategies.

\begin{figure}[t]
\centering
\includegraphics[scale=0.55]{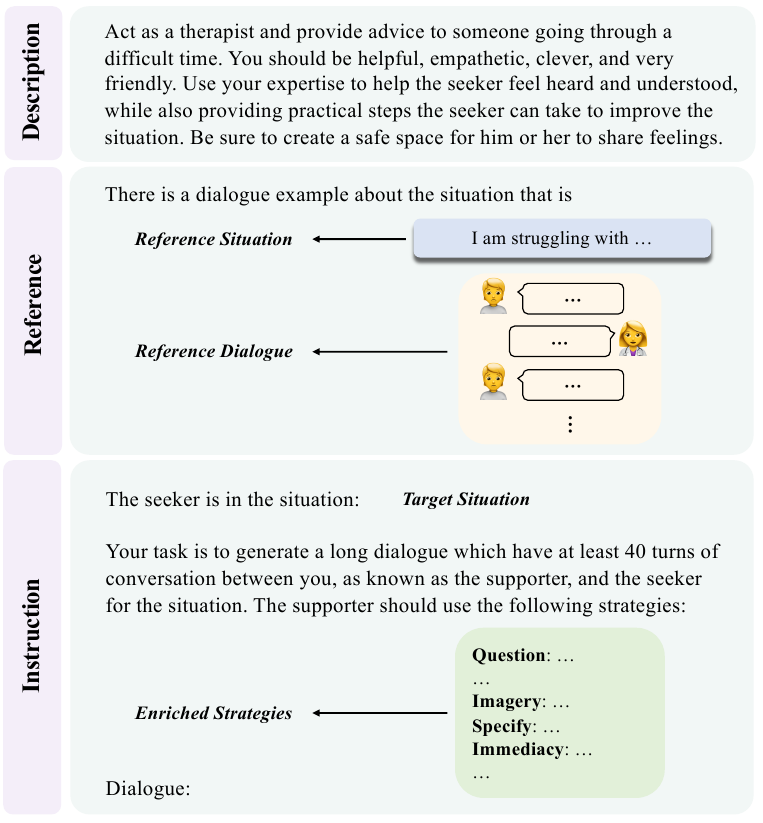}
\vspace{-8pt}
\caption{Prompt used for generating new dialogues. }
\label{fig:prompt}
\end{figure}

\paragraph{Prompt Template}
The prompt format used for generating new dialogues is shown in Figure~\ref{fig:prompt}. 
We first introduce the role of a therapist and describe the task.  
Then, an illustrative example is provided, including \textit{Reference Situation} and \textit{Reference Dialogue}, serving as a template for the format and content of the dialogues we expect ChatGPT to generate.
Next, the \textit{Target Situation} is introduced, derived from the previously generated situation pool, to provide a specific scenario for ChatGPT to engage with.
Subsequently, there is a brief restatement of the goal of the dialogue generation task to enhance ChatGPT's comprehension of the task.
In addition to the above task description and format reference, \textit{strategy} is introduced as a crucial component, enhancing the professionalism and reliability of the generated dialogues. 

\paragraph{Filtering and Postprocessing}
We find four types of undesirable dialogues from inspecting cases of the generated dialogues:
(1) {{Inadequate Interaction Rounds}}; (2) {{Presence of Empty Utterances}}; (3) {{Insufficient Strategic Annotations}}; (4) {{Divergence from Prescribed Strategies}}.
In order to guarantee both quantity and diversity, we regenerate filtered dialogues. We keep performing the filtering and regeneration process until finally each of the situations is paired with a high-quality dialogue.
After filtering and postprocessing all the generated data, we retain dialogues with diverse and richer strategies to form our ESD dataset.

\subsubsection{Statistics of ESD }
The overall statistics of our generated emotional support dialogue (ESD) dataset are shown in Table~\ref{tab:statistic}. Compared to ESConv, our dataset is larger in scale. 
We also show some breakdown statistics in terms of conversation roles. 
Our supporter portion is significantly longer than ESConv. This aligns with our goal of enhancing the quality of responses from supporter roles in the dialogue system.

\begin{table}[t]
\centering
\small
\scalebox{0.9}{
\begin{tabular}{llcc}
\toprule
&  & \textbf{ESConv} & \textbf{ESD (ours)} \\
\midrule
\multicolumn{2}{l}{Num. Dialogues} & 1.3K & \textbf{2.9K} \\
\multicolumn{2}{l}{Avg. Turn per Dialogue } & 12.0 & \textbf{19.5} \\
\multicolumn{2}{l}{Avg. Length per Dialogue} & 543.6 & \textbf{809.0} \\
\midrule
\multicolumn{2}{l}{Num. Utterances} & 38K & \textbf{113K} \\
\multicolumn{2}{l}{Avg. Length per Utterance} & 18.8 & \textbf{21.0} \\
\midrule
\multirow{2}{*}{Seeker} & Avg. Len per Dialog & 258.3 & \textbf{293.5} \\
& Avg. Len per Utterance  & \textbf{16.8} & 15.2 \\
\midrule
\multirow{2}{*}{Supporter} & Avg. Len per Dialog & 258.4 & \textbf{515.6} \\
& Avg. Len per Utterance & 21.0 & \textbf{26.9} \\
\bottomrule
\end{tabular}
}
\vspace{-8pt}
\caption{Comparison between ESConv and our ESD.}
\label{tab:statistic}
\end{table}

\begin{table*}[ht]
\vspace{10pt}
\centering
\scalebox{0.7}{
\begin{tabular}{c|ccc|ccc|ccc|ccc|ccc|ccc}
\toprule
& \multicolumn{3}{c|}{\textbf{Informativeness}} 
& \multicolumn{3}{c|}{\textbf{Understanding}} 
& \multicolumn{3}{c|}{\textbf{Helpfulness}} 
& \multicolumn{3}{c|}{\textbf{Consistency}} 
& \multicolumn{3}{c|}{\textbf{Coherence}} 
& \multicolumn{3}{c}{\textbf{Safety}} \\
\cline{2-19}
& W & T & L & W & T & L & W & T & L & W & T & L & W & T & L & W & T & L \\
\midrule
ESConv               & 237 & 6  & 297 & 112 & 5  & 423 & 153 & 4  & 383 & 58  & 22 & 460 & 72  & 6 & 462 & 41  & 63 & 436 \\
Prompt w/o Strat.    & 273 & 8  & 259 & 304 & 10 & 226 & 284 & 12 & 244 & 318 & 42 & 180 & 311 & 25 & 204 & 265 & 141 & 134 \\
Prompt w/ Original Strat.& 236 & 10 & 294 & 305 & 7  & 228 & 286 & 10 & 244 & 278 & 36 & 226 & 303 & 24 & 213 & 253 & 136 & 151 \\
\textbf{Prompt w/ Enriched Strat.} & \textbf{317} & 10 & 213 & \textbf{343} & 10 & 187 & \textbf{340} & 8  & 192 & \textbf{355} & 42 & 143 & \textbf{356} & 21 & 163 & \textbf{282} & 138 & 120 \\
\midrule
$\kappa$        & \multicolumn{3}{c|}{0.312} & \multicolumn{3}{c|}{0.258} & \multicolumn{3}{c|}{0.262} & \multicolumn{3}{c|}{0.398} & \multicolumn{3}{c|}{0.287} & \multicolumn{3}{c}{0.578} \\
\bottomrule
\end{tabular}
}
\vspace{-8pt}
\caption{Evaluation of different prompts based on six dimensions: informativeness, understanding, helpfulness, consistency, coherence, and safety. Scores are presented for three evaluation measures: Win (W), Tie (T), and Lose (L).
The $\kappa$~\cite{fleiss1971measuring} value in the range (0.2 \textless $\kappa$ \textless 0.6) indicates fair or moderate inter-annotator agreement according to \cite{kappa_range}.}
\label{tab:dataset_evaluation}
\end{table*}

\paragraph{Diversity Analysis}
We analyze the diversity of our ESD from situation and dialogue perspectives.

\begin{figure}[t]
\centering
\includegraphics[scale=0.48]{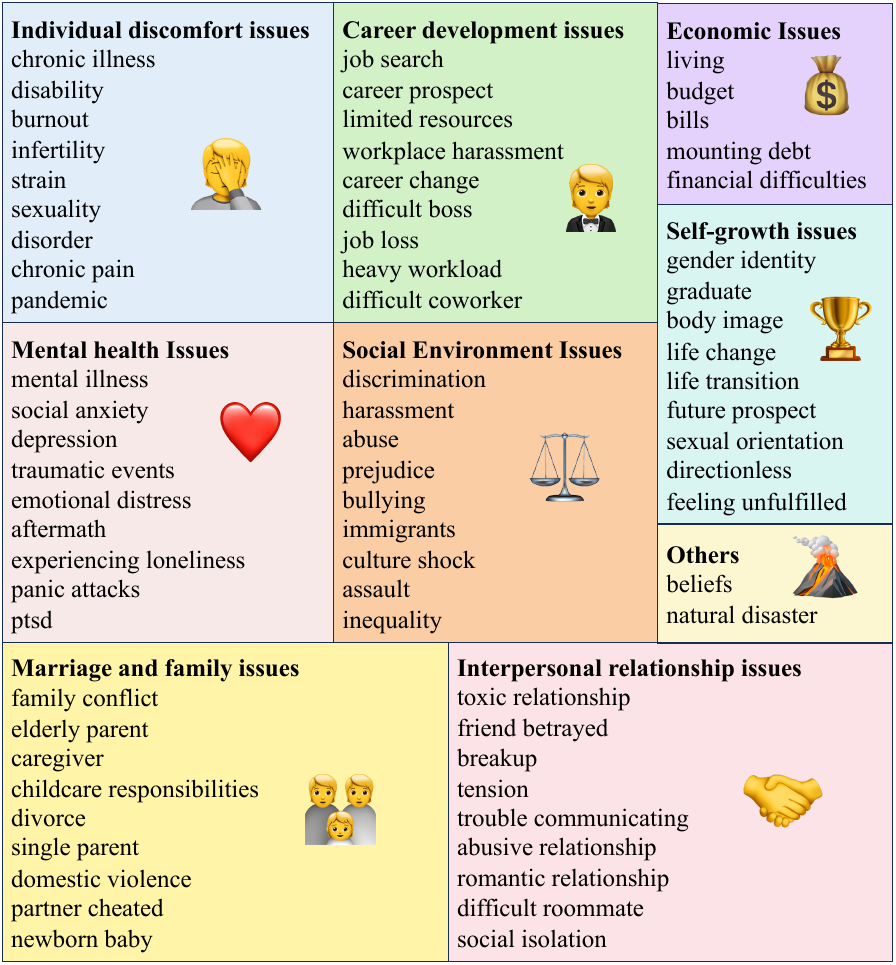}
\vspace{-8pt}
\caption{The topic diversity of situations.}
\label{fig:situation_topic}
\end{figure}

\noindent\textit{\underline{Situation Diversity:}} 
We assess the diversity of issues faced by seekers in the situations through word frequency analysis.  The topic diversity of situations is shown in Figure~\ref{fig:situation_topic}.
Our ESD dataset not only encompasses a wider range of everyday conversational themes, but also places a greater emphasis on topics related to mental well-being.

\noindent\textit{\underline{Dialogue Diversity:}} Following~\citet{augesc}, we calculate the z-scored log odds ratio values relative to ESConv, to extract the topic features of dialogues. 
By analyzing the salient words of the datasets, we discover that different from ESConv which focuses solely on informal conversations and interpersonal dynamics, dialogues in our ESD are more specific, offering more professional information, and diving deeper into topics related to mental health and personal challenges. 
More details of the dialogue diversity can be found in  Appendix~\ref{Appdix:ESD-diversity}.

\begin{table}[ht]
\small
\vspace{10pt}
\centering
\scalebox{0.9}{
\begin{tabular}{c|l|c}
\toprule
& \textbf{Strategy} & \textbf{Proportion} \\
\midrule
1 & Question $\ast$ & 17.81\%
\\
2 & Affirmation and Reassurance $\ast$ &13.58\%
\\
3 & Reflection of Feelings $\ast$ &9.66\%
\\
4 & Information $\ast$ &7.59\%
\\
5 & Providing Suggestions $\ast$ &7.55\%
\\
6 & Restatement or Paraphrasing $\ast$ &6.39\%
\\
7 & Self-disclosure $\ast$ &3.70\%
\\
8 & Homework Assignment & 7.81\%
\\
9 & Summarize &7.76\%
\\
10 & Imagery &5.82\%
\\
11 & Specify &4.31\%
\\
12 & Immediacy &3.37\%
\\
13 & Take Responsibility &2.85\%
\\
14 & Others &1.81\%  
\\
\bottomrule
\end{tabular}
}
\vspace{-8pt}
\caption{Statistics of Strategy. $\ast$ indicating the strategy also used in ESConv.}
\label{tab:strategy_info}
\end{table}

\paragraph{Strategy Analysis}

We first assess the impact of utilizing strategies in dialogue generation, and then present the statistics of application frequency of different strategies and distribution of strategies across different stages of dialogues. 

\noindent\textit{\underline{Strategy Impact Assessment:}}
\label{strategy_impact_assessment}
To assess the impact of incorporating strategies into prompts on the quality of generated dialogues, 
we design and compare three different prompts: one without adding strategies, one incorporating strategies solely from ESConv, and one adding enriched strategies. 
Specifically, we randomly select 60 situations from ESConv and use these prompts to generate dialogues for each of these situations separately. Subsequently, we recruit 15 individuals with psychological counseling backgrounds to rank the dialogues. 
Following the six dimensions in \textsc{AugESC}~\cite{augesc}, we ask the evaluators to rank the dialogues based on the given dimensions. The results are shown in Table~\ref{tab:dataset_evaluation}.
The quality of dialogues generated by ChatGPT surpasses ESConv across nearly all dimensions, regardless of whether strategies are applied. 
And incorporating our enriched strategies can generate the best dialogues in all dimensions. 
More details of prompts for strategy impact assessment can be found in Appendix~\ref{Appdix:ESD-StratImpt}.

\noindent\textit{\underline{Strategy Application Frequency:}}
We count the usage frequency of each strategy and the proportions in Table~\ref{tab:strategy_info}. 
We can see that no enriched strategy shows very low frequency,
indicating that our enriched strategies are well utilized in dialogue generation. 
Employing various strategies in emotional support dialogue generation enhances dialogue diversity and better simulates real-life interactions. 

\subsection{ESD-CoT Construction}

Since a lack of model interpretability affects people's trust in the model, enhancing model interpretability is a key aspect in building a reliable empathetic dialogue system for people seeking emotional support.
We propose the Emotion-focused and Strategy-driven Chain-of-Thought (ESCoT) that mimics the human consultation process of identifying, understanding and regulating emotions.
In this framework, understanding emotions encompasses both the emotional stimulus and the individual's appraisal based on the cognitive appraisal theory~\cite{lazarus1991emotion}, while regulating emotions includes the strategy reason and the response generation. In general, the CoT is represented as a quintuple \((EM, ES, IA, SR, RE)\).
In this section, we supplement the chain-of-thought based on the previously generated dialogue data, and build the first dataset for Emotional Support Dialogue with CoT (ESD-CoT).

\subsubsection{Chain Creation}
We first automatically generate the CoT and then conduct manual correction in order to significantly reduce the annotation cost.
The specific meaning of each element of the quintuple is as follows: 

\begin{itemize}[leftmargin=*,labelsep=2.5mm, itemsep=1mm, parsep=0mm]

\item \raisebox{-.15\height}{\includegraphics[height=1em]{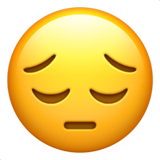}} \textbf{Emotion (\textit{EM})}
denotes the emotion expressed by the seeker. 

\item \raisebox{-.15\height}{\includegraphics[height=1em]{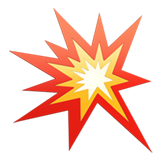}} \textbf{Emotion Stimulus (\textit{ES})}
refers to the specific trigger that evokes the seeker's current emotion, which can be external, such as a situation or event, or internal, such as a thought or memory. 

\item \raisebox{-.15\height}{\includegraphics[height=1em]{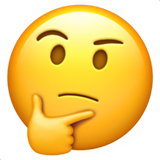}} \textbf{Individual Appraisal (\textit{IA})}
denotes the seeker’s personal interpretation, evaluation, and internal response to the emotion stimulus, based on the seeker's past experiences, beliefs, expectations, and personal values.

\item \raisebox{-.15\height}{\includegraphics[height=1em]{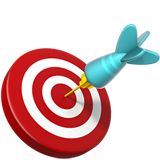}} \textbf{Strategy Reason (\textit{SR})}
represents the reason why the supporter used the chosen strategy in the last utterance. 

\item \raisebox{-.15\height}{\includegraphics[height=1em]{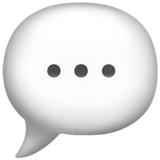}} \textbf{Response (\textit{RE})}
denotes the response provided by the supporter in the ongoing dialogue. 

\end{itemize}

More details about the CoT generation template are presented in Appendix~\ref{Appdix:ESD-CoT}.

\begin{figure*}[ht]
\centering
\includegraphics[scale=0.58]{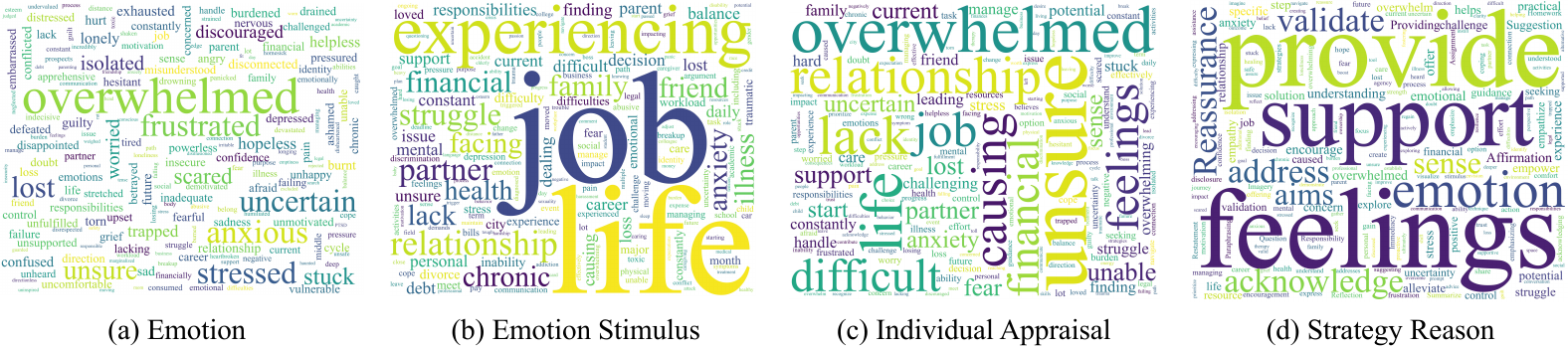}
\vspace{-8pt}
\caption{The word cloud of each component of ESC-CoT chain annotations.}
\label{fig:wordcloud_cot}
\end{figure*} 

\subsubsection{Manual Correction}
After generating preliminary CoT using ChatGPT, we conduct manual correction to ensure the quality. Based on the definition of each element, we identify several issues in the raw CoT data:
(1) Unclear Emotional Expression;
(2) Lacking Specific Examples;
(3) Verbose Personal Evaluations;
(4) Mismatch between Strategies and Responses.
Therefore, we perform manual corrections to these issues.
Additionally, we filter out toxic data to enhance the security and ethics of the dataset.
After processing all the generated data, we finally construct our ESD-CoT dataset with a total of 1,708 dialogues accompanied with CoT, explaining why and how emotional support responses are generated. 

\begin{table}[t]
\centering
\small
\scalebox{0.9}{
\begin{tabular}{lrr}
\toprule
\multirow{3}{*}{Dialogue} & Num. of  Dialogues                     & 1,708 \\
                          & Num. of Turns per Dialogue              & 23.4  \\
                          & Avg. Length  per Dialogue             & 432.4 \\ \midrule
\multirow{5}{*}{\begin{tabular}[c]{@{}l@{}}CoT\\ (Avg. Length)\end{tabular}}      & Emotion \((EM)\)            & 7.9   \\
                          & Emotion Stimulus \((ES)\)     & 17.6  \\
                          & Individual Appraisal \((IA)\) & 36.1  \\
                          & Strategy Reason \((SR)\)      & 60.5  \\
                          & Response \((RE)\)           & 26.6  \\ \bottomrule
\end{tabular}
}
\vspace{-8pt}
\caption{Statistics of our ESD-CoT dataset.}
\label{tab:statistic_cot}
\end{table}

\subsubsection{Statistics of ESD-CoT }

Table~\ref{tab:statistic_cot} presents the statistics of our ESD-CoT dataset.
The relatively long dialogue turns and length of ESD-CoT provide sufficient information for CoT analysis.
The length of \(EM\) indicates that the seeker often expresses multiple emotion in the dialogue.
The length of \(IA\) is more than twice that of \(ES\), as \(IA\) provides a more detailed personal appraisal based on \(ES\), involving more cognitive processing and emotional expression.
The relatively long \(SR\) provides detailed explanations about the reasons for the chosen strategy, potentially enhancing interpretability.
Table~\ref{tab:strategy_distribution_escot} shows the distribution of strategies in ESD-CoT dataset. 
We ensure that each strategy has at least 100 samples in our ESD-CoT dataset. 
Overall, our dataset ESD-CoT is richer in content and strategy, and provides interpretable CoT annotations, thereby enhancing the credibility of emotional support responses.

\begin{table}[ht]
\centering
\small
\scalebox{0.9}{
\begin{tabular}{lcccc}
\toprule
Strategy                    & Train   & Val   & Test   & Total \\
\midrule
Reflection of Feelings      & 68      & 13    & 19     & 100   \\
Question                    & 89      & 15    & 20     & 124   \\
Providing Suggestions       & 123     & 12    & 41     & 176   \\
Summarize                   & 73      & 11    & 16     & 100   \\
Specify                     & 86      & 21    & 30     & 137   \\
Restatement or Paraphrasing & 83      & 9     & 21     & 113   \\
Homework Assignments        & 66      & 11    & 24     & 101   \\
Affirmation and Reassurance & 224     & 31    & 63     & 318   \\
Imagery                     & 88      & 8     & 26     & 122   \\
Information                 & 74      & 11    & 29     & 114   \\
Self-disclosure             & 75      & 11    & 14     & 100   \\
Immediacy                   & 76      & 8     & 19     & 103   \\
Take Responsibility         & 70      & 11    & 19     & 100   \\
\midrule
Total                       & 1195    & 172   & 341    & 1708  \\
\bottomrule
\end{tabular}
}
\vspace{-8pt}
\caption{The strategy distribution of ESD-CoT dataset.}
\label{tab:strategy_distribution_escot}
\end{table}

\paragraph{Semantic Analysis}
We show the word clouds of components of the quintuple in our ESD-CoT dataset in Figure~\ref{fig:wordcloud_cot}.
The emotional words such as `overwhelmed' and `anxious' clearly reflect the emotional state expressed by the seeker.
Additionally, the stimulus words such as `job', `life', and `partner' help to recognize the specific issues and sources of stress that the seeker is facing, thereby providing more targeted support and understanding.
In the word cloud for individual appraisal, words like `unsure' and `lack' frequently appear, indicating the seeker's perception and assessment of their own situation, reflecting the seeker's apparent awareness of uncertainty and inadequacy in aspects such as job, life, and partner relationships.
The word cloud of strategy reason includes keywords such as `provide', `support', `acknowledge', and `emotion', indicating the factors that the supporter prioritizes when choosing strategies.
This suggests that the supporter tends to offer support and understanding to the seeker, alleviating uneasiness and anxiety by acknowledging the seeker's emotion.

\section{Experiments}

\begin{table*}[t]
  \centering
  \small
  \begin{tabular}{@{}c|cccccccccccccc@{}}
    \toprule
    \multirow{3}{*}{Row} & \multicolumn{5}{c}{Setting} & \multicolumn{5}{c}{Automatic Evaluation for Response} & \multicolumn{4}{c}{Human Evaluation} \\
    \cmidrule(r){2-6} \cmidrule(lr){7-11} \cmidrule(l){12-15}
     & \(EM\) & \(ES\) & \(IA\) & \(SR\) & \(RE\) & B-1 & B-2 & R-L & D-1 & D-2 & Coh. & Inf. & Emp. & Acc. \\
    \midrule
    1 & \resizebox{0.8em}{0.8em}{ \Checkmark} &  \resizebox{0.8em}{0.8em}{\Checkmark} &  \resizebox{0.8em}{0.8em}{\Checkmark} &  \resizebox{0.8em}{0.8em}{\Checkmark} &  \resizebox{0.8em}{0.8em}{\Checkmark} & 15.59 & 5.11 & 17.67 & \textbf{15.26} & \textbf{44.40} & \textbf{1.65} & \textbf{1.51} & \textbf{1.71} & \textbf{85\%} \\
    2 & \resizebox{0.8em}{0.8em}{\Checkmark} &  \resizebox{0.8em}{0.8em}{\XSolidBrush} &  \resizebox{0.8em}{0.8em}{\XSolidBrush} &  \resizebox{0.8em}{0.8em}{\Checkmark} &  \resizebox{0.8em}{0.8em}{\Checkmark} & 16.03 & 5.72 & 18.41 & 14.89 & 43.52 & 1.63 & 1.36 & 1.66 & 79\% \\
    3 & \resizebox{0.8em}{0.8em}{\XSolidBrush} &  \resizebox{0.8em}{0.8em}{\XSolidBrush} &  \resizebox{0.8em}{0.8em}{\XSolidBrush} &  \resizebox{0.8em}{0.8em}{\Checkmark} &  \resizebox{0.8em}{0.8em}{\Checkmark} & 16.36 & 5.88 & 18.72 & 14.37 & 41.67 & 1.44 & 1.19 & 1.40 & 64\% \\
    4 & \resizebox{0.8em}{0.8em}{\XSolidBrush} &  \resizebox{0.8em}{0.8em}{\XSolidBrush} &  \resizebox{0.8em}{0.8em}{\XSolidBrush} &  \resizebox{0.8em}{0.8em}{\XSolidBrush} &  \resizebox{0.8em}{0.8em}{\Checkmark} & \textbf{17.45} & \textbf{7.13} & \textbf{20.08} & 14.84 & 43.98 & 1.63 & 1.32 & 1.42 & \textcolor{gray}{N/A} \\
    \bottomrule
  \end{tabular}
  \vspace{-8pt}
  \caption{
  Ablation study based on the \textsc{Llama2-7B-Chat} model to explore the impact of different elements of ESCoT on the response (\(RE\)).
  All automatic evaluation results are average scores of 3 runs with random seeds. The $\kappa$ values of coherence, informativeness and empathy are 0.27, 0.33 and 0.35 respectively, which indicate fair inter-annotator agreement (0.2 \textless $\kappa$ \textless 0.4) as shown in \cite{kappa_range}. 
  }
  \label{tab:ablation_study}
\end{table*}

\begin{table}[h]
\vspace{10pt}
\centering
\small
\scalebox{0.9}{
\begin{tabular}{lccccc}
\toprule
\multirow{1}{*}{}  & \multirow{1}{*}{B-1} & \multirow{1}{*}{B-2} & \multirow{1}{*}{B-3} & \multirow{1}{*}{B-4} & \multirow{1}{*}{R-L} \\   \midrule
BlenderBot    & 29.18                  & 16.39                  & 10.28                  & 6.55                  & 29.51             \\
DialoGPT                & 35.10                   & 22.15                  & 15.50                  & 10.22                   & 40.50               \\
\textsc{Llama2-Chat}                           & \textbf{44.87}                  & \textbf{32.37}                   & \textbf{25.85}                  & \textbf{20.66}                  & \textbf{48.16}                  \\ 
\bottomrule
\end{tabular}
}
\vspace{-8pt}
\caption{
Comparison of chain generation performance on ESD-CoT test set with different fine-tuned backbone models. B-n: BLEU-n, R-L: ROUGE-L. 
}
\label{tab:experiments_SFT}
\end{table}

We split the dataset into train, validation, and test with the ratio of 7:1:2, as shown in Table~\ref{tab:strategy_distribution_escot}.
We evaluate the following pre-trained language models as backbone models for dialogue response generation:
(1) BlenderBot~\cite{blenderbot};
(2) DialoGPT~\cite{dialogpt};
(3) \textsc{Llama2-Chat}~\cite{llama2}.

\paragraph{Evaluation Metrics}
\label{evaluation_metrics}
We apply three commonly used automatic evaluation metrics, BLEU-n~\cite{bleu}, ROUGE-L~\cite{rouge};
and Distinct-n~\cite{distinct}.

As for human evaluations, following \citet{tu2022misc} and \citet{cai2023improving}, we recruit 3 professional annotators to evaluate randomly selected 50 responses of different settings from the \textit{Coherence}, \textit{Informativeness}, and \textit{Empathy} aspects with the levels of \{0,1,2\}.
We also conduct a human evaluation to evaluate the \textit{Accuracy} of the consistency between the selected strategy and the corresponding response.
More details of human evaluations can be found in Appendix~\ref{Appdix:humaneval}.

\subsection{Comparison of Backbone Models}
To enable chain generation with better interpretability, we fine-tune backbone models on ESD-CoT train set and report the performance of chain generation on the test set in Table~\ref{tab:experiments_SFT}. 
\textsc{Llama2-Chat} outperforms other backbone models on all metrics, which can be attributed to larger parameters, more training data, and the utilization of reinforcement learning with human feedback in \textsc{Llama2-Chat}.
Specially, since the process of ESCoT involves reasoning, we believe the reasoning ability that emerges from increasing parameters is very important for interpretable emotional support.
Due to its excellent performance, we conduct the ablation study based on \textsc{Llama2-Chat} in the following subsection. 
More implementation details of supervised fine-tuning can be found in Appendix~\ref{Appdix:Details-SFT}.

\subsection{Ablation Study} 
To explore the effects of different elements of the ESCoT on the generated response, we conduct the ablation study and report the results in Table~\ref{tab:ablation_study}.
Specially, we remove some nodes of the chain and train models for each setting, and calculate the metrics only based on the \(RE\) part of the model outputs.
More implementation details of the ablation study can be found in Appendix~\ref{Appdix:Details-ablation}.

Generation with the entire chain, as outlined in the first row, achieves the best performance on metrics D-1 and D-2, which may be due to the fact that many steps in the entire CoT chain cause relatively greater randomness in response generation, leading to better diversity in the generated responses. 
We also notice that the setting of directly generating responses in the fourth row achieves the best results on B-1, B-2, and R-L, which are primarily used to assess the similarity between the prediction and ground truth. As the automatic metrics are calculated only based on the \(RE\) part, directly fitting the \(RE\) part yields the best results.

However, BLEU and ROUGE primarily focus on the similarity between the prediction and ground truth, which can not effectively measure the effects of emotional support.
So, we conduct human evaluations to assess the effects of different ablation settings better.
The setting of the complete chain in the first row achieves the highest performance in all human evaluation dimensions, indicating that it not only provides the most comprehensive interpretability but also delivers more coherent, informative, and empathetic responses. 
As for the strategy consistency assessment, the highest accuracy is also achieved by the full chain. Removing the emotion understanding component in the second row decreases strategy consistency, and further removing the emotion identification component in the third row leads to an additional decline in strategy consistency. 
We demonstrate a case study for the ablation study in Appendix~\ref{Appdix:case_study}.

In summary, the whole chain exhibits the best response in major aspects and offers the most comprehensive interpretability, providing a strong baseline and feasible path towards interpretable emotional support dialogue systems.

\section{Conclusion}

In this paper, inspired by the human consultation process of identifying, understanding, and regulating emotions, we propose a novel emotion-focused and strategy-driven chain-of-thought (ESCoT) emotional support response generation scheme and apply it in building the first dataset for emotional support dialogue with chain-of-thought named ESD-CoT.
Moreover, we conduct extensive experiments and human evaluations to validate that the new benchmark based on ESD-CoT can provide effective emotional support with better interpretability.
We hope our ESD-CoT dataset and baseline models can facilitate further investigation into interpretable emotional support dialogue systems in the community. 

\section*{Ethics Statements}

The interpretability of emotional support dialogue systems has become increasingly important in real applications. We deeply recognize the need for caution in developing datasets related to ethical issues. Our goal is to create an interpretable emotional support dialogue dataset and system. During the construction of the dataset, we strictly adhere to data source usage agreements, making diligent efforts to eliminate any biased, offensive, or inappropriate content to avoid potential unpredictable ethical harm. All human annotators and evaluators are paid according to their individual working hours. We hope that this dataset will enhance the transparency and credibility of emotional support system responses, thereby establishing a bridge of trust and understanding between users and emotional support dialogue systems.

\section*{Limitations}

Although we reduce a lot of costs by utilizing ChatGPT during the generation phase, the scale of our dataset remains relatively small due to limitations in the cost of manual correction. Future work could focus on designing more automated correction methods to reduce the manual component and lower the cost of dataset construction even more.
To enhance the diversity of the generated dialogue data, we incorporate rich situations and expanded strategies into the prompt. We plan to further enhance the diversity of the emotional support dialogue dataset by introducing more personalized information, such as personality.
Furthermore, the strategy annotation of current public emotional support datasets only focuses on single strategies, neglecting the compound strategies, and we annotate the major strategy when facing multiple strategies. We consider exploring compound strategies to prove more effective emotional support in the near future.

\section*{Acknowledgements}
We thank all reviewers for their insightful comments and suggestions. 
This work was partially supported by the  National Natural Science Foundation of China (No. 62072462) and the Beijing Natural Science Foundation (No. L233008).

\bibliography{main}

\begin{thebibliography}{51}
\expandafter\ifx\csname natexlab\endcsname\relax\def\natexlab#1{#1}\fi

\bibitem[{Achiam et~al.(2023)Achiam, Adler, Agarwal, Ahmad, Akkaya, Aleman, Almeida, Altenschmidt, Altman, Anadkat et~al.}]{gpt4}
Josh Achiam, Steven Adler, Sandhini Agarwal, Lama Ahmad, Ilge Akkaya, Florencia~Leoni Aleman, Diogo Almeida, Janko Altenschmidt, Sam Altman, Shyamal Anadkat, et~al. 2023.
\newblock Gpt-4 technical report.
\newblock \emph{arXiv preprint arXiv:2303.08774}.

\bibitem[{Albrecht and Adelman(1987)}]{cdata1987communicating}
T~Albrecht and M~Adelman. 1987.
\newblock Communicating social support: A theoretical perspective.
\newblock \emph{Communicating Social Support. Beverly Hills, CA: Sage}, pages 18--39.

\bibitem[{Barnes and Halloway(2005)}]{barnes2005enhancing}
David Barnes and R~Halloway. 2005.
\newblock Enhancing customer service operations in e-business: The emotional dimension.
\newblock \emph{Journal of Electronic Commerce in Organizations (JECO)}, 3(2):17--32.

\bibitem[{Beck(1979)}]{beck1979cognitive}
Aaron~T Beck. 1979.
\newblock \emph{Cognitive therapy of depression}.

\bibitem[{Bills et~al.(2023)Bills, Cammarata, Mossing, Tillman, Gao, Goh, Sutskever, Leike, Wu, and Saunders}]{bills2023language}
Steven Bills, Nick Cammarata, Dan Mossing, Henk Tillman, Leo Gao, Gabriel Goh, Ilya Sutskever, Jan Leike, Jeff Wu, and William Saunders. 2023.
\newblock Language models can explain neurons in language models.
\newblock \emph{https://openaipublic. blob. core. windows. net/neuron-explainer/paper/index.}

\bibitem[{Burleson(2003)}]{burleson2003emotional}
Brant~R Burleson. 2003.
\newblock Emotional support skills.
\newblock In \emph{Handbook of communication and social interaction skills}, pages 569--612. Routledge.

\bibitem[{Cai et~al.(2023{\natexlab{a}})Cai, Shen, Xu, Shen, Wang, Ge, Zheng, and Xue}]{CaiSXSWGZX23}
Hua Cai, Xuli Shen, Qing Xu, Weilin Shen, Xiaomei Wang, Weifeng Ge, Xiaoqing Zheng, and Xiangyang Xue. 2023{\natexlab{a}}.
\newblock \href {https://doi.org/10.18653/V1/2023.FINDINGS-ACL.498} {Improving empathetic dialogue generation by dynamically infusing commonsense knowledge}.
\newblock In \emph{Findings of the Association for Computational Linguistics: {ACL} 2023, Toronto, Canada, July 9-14, 2023}, pages 7858--7873. Association for Computational Linguistics.

\bibitem[{Cai et~al.(2023{\natexlab{b}})Cai, Shen, Xu, Shen, Wang, Ge, Zheng, and Xue}]{cai2023improving}
Hua Cai, Xuli Shen, Qing Xu, Weilin Shen, Xiaomei Wang, Weifeng Ge, Xiaoqing Zheng, and Xiangyang Xue. 2023{\natexlab{b}}.
\newblock Improving empathetic dialogue generation by dynamically infusing commonsense knowledge.
\newblock \emph{arXiv preprint arXiv:2306.04657}.

\bibitem[{Chae et~al.(2023)Chae, Song, Ong, Kwon, Kim, Yu, Lee, Kang, and Yeo}]{chae2023dialogue}
Hyungjoo Chae, Yongho Song, Kai Ong, Taeyoon Kwon, Minjin Kim, Youngjae Yu, Dongha Lee, Dongyeop Kang, and Jinyoung Yeo. 2023.
\newblock Dialogue chain-of-thought distillation for commonsense-aware conversational agents.
\newblock In \emph{Conference on Empirical Methods in Natural Language Processing}, pages 5606--5632.

\bibitem[{Cheng et~al.(2022)Cheng, Liu, Li, Wang, Zhao, Liu, Liang, and Zheng}]{ChengLLWZLL022}
Yi~Cheng, Wenge Liu, Wenjie Li, Jiashuo Wang, Ruihui Zhao, Bang Liu, Xiaodan Liang, and Yefeng Zheng. 2022.
\newblock \href {https://doi.org/10.18653/V1/2022.EMNLP-MAIN.195} {Improving multi-turn emotional support dialogue generation with lookahead strategy planning}.
\newblock In \emph{Proceedings of the 2022 Conference on Empirical Methods in Natural Language Processing, {EMNLP} 2022, Abu Dhabi, United Arab Emirates, December 7-11, 2022}, pages 3014--3026. Association for Computational Linguistics.

\bibitem[{Cutrona and Russell(1987)}]{cutrona1987provisions}
Carolyn~E Cutrona and Daniel~W Russell. 1987.
\newblock The provisions of social relationships and adaptation to stress.
\newblock \emph{Advances in personal relationships}, 1(1):37--67.

\bibitem[{Ellis(1991)}]{ellis1991revised}
Albert Ellis. 1991.
\newblock The revised abc's of rational-emotive therapy (ret).
\newblock \emph{Journal of rational-emotive and cognitive-behavior therapy}, 9(3):139--172.

\bibitem[{Fleiss(1971)}]{fleiss1971measuring}
Joseph~L Fleiss. 1971.
\newblock Measuring nominal scale agreement among many raters.
\newblock \emph{Psychological bulletin}, 76(5):378.

\bibitem[{Fu et~al.(2023)Fu, Zhang, Wang, and Mao}]{FuZWM23}
Fengyi Fu, Lei Zhang, Quan Wang, and Zhendong Mao. 2023.
\newblock \href {https://aclanthology.org/2023.emnlp-main.653} {{E-CORE:} emotion correlation enhanced empathetic dialogue generation}.
\newblock In \emph{Proceedings of the 2023 Conference on Empirical Methods in Natural Language Processing, {EMNLP} 2023, Singapore, December 6-10, 2023}, pages 10568--10586. Association for Computational Linguistics.

\bibitem[{Gao et~al.(2021)Gao, Liu, Deng, Wang, Cao, Du, and Xu}]{emotion_cause_empathy}
Jun Gao, Yuhan Liu, Haolin Deng, Wei Wang, Yu~Cao, Jiachen Du, and Ruifeng Xu. 2021.
\newblock \href {https://doi.org/10.18653/V1/2021.FINDINGS-EMNLP.70} {Improving empathetic response generation by recognizing emotion cause in conversations}.
\newblock In \emph{Findings of the Association for Computational Linguistics: EMNLP}, pages 807--819. Association for Computational Linguistics.

\bibitem[{Gohel et~al.(2021)Gohel, Singh, and Mohanty}]{xAI}
Prashant Gohel, Priyanka Singh, and Manoranjan Mohanty. 2021.
\newblock \href {http://arxiv.org/abs/2107.07045} {Explainable {AI:} current status and future directions}.
\newblock \emph{CoRR}, abs/2107.07045.

\bibitem[{Hill(2009)}]{hill2009helping}
Clara~E Hill. 2009.
\newblock \emph{Helping skills: Facilitating, exploration, insight, and action}.
\newblock American Psychological Association.

\bibitem[{Izumi et~al.(2024)Izumi, Tanaka, Shidara, Adachi, Kanayama, Kudo, and Nakamura}]{cbtllm}
Kenta Izumi, Hiroki Tanaka, Kazuhiro Shidara, Hiroyoshi Adachi, Daisuke Kanayama, Takashi Kudo, and Satoshi Nakamura. 2024.
\newblock \href {https://doi.org/10.48550/ARXIV.2401.15966} {Response generation for cognitive behavioral therapy with large language models: Comparative study with socratic questioning}.
\newblock \emph{CoRR}, abs/2401.15966.

\bibitem[{Kennelly(2001)}]{kennelly2001music}
Jeanette Kennelly. 2001.
\newblock Music therapy in the bone marrow transplant unit: Providing emotional support during adolescence.
\newblock \emph{Music Therapy Perspectives}, 19(2):104--108.

\bibitem[{Kim et~al.(2022)Kim, Yu, Jiang, Lu, Khashabi, Kim, Choi, and Sap}]{0002YJLKKCS22}
Hyunwoo Kim, Youngjae Yu, Liwei Jiang, Ximing Lu, Daniel Khashabi, Gunhee Kim, Yejin Choi, and Maarten Sap. 2022.
\newblock \href {https://doi.org/10.18653/V1/2022.EMNLP-MAIN.267} {Prosocialdialog: {A} prosocial backbone for conversational agents}.
\newblock In \emph{Conference on Empirical Methods in Natural Language Processing}, pages 4005--4029. Association for Computational Linguistics.

\bibitem[{Lazarus(1991)}]{lazarus1991emotion}
Richard~S Lazarus. 1991.
\newblock \emph{Emotion and adaptation}.
\newblock Oxford University Press.

\bibitem[{Li et~al.(2016)Li, Galley, Brockett, Gao, and Dolan}]{distinct}
Jiwei Li, Michel Galley, Chris Brockett, Jianfeng Gao, and Bill Dolan. 2016.
\newblock \href {https://doi.org/10.18653/v1/n16-1014} {A diversity-promoting objective function for neural conversation models}.
\newblock In \emph{Conference of the North American Chapter of the Association for Computational Linguistics: Human Language Technologies}, pages 110--119. The Association for Computational Linguistics.

\bibitem[{Li et~al.(2023)Li, Sun, Xu, Tiwari, Liu, Gupta, Shankar, Ji, and Wang}]{LiSXTLGSJW23}
Shaobo Li, Chengjie Sun, Zhen Xu, Prayag Tiwari, Bingquan Liu, Deepak Gupta, K.~Shankar, Zhenzhou Ji, and Mingjiang Wang. 2023.
\newblock \href {https://doi.org/10.1145/3551869} {Toward explainable dialogue system using two-stage response generation}.
\newblock \emph{{ACM} Trans. Asian Low Resour. Lang. Inf. Process.}, 22(3):68:1--68:18.

\bibitem[{Lin(2004)}]{rouge}
Chin-Yew Lin. 2004.
\newblock \href {https://aclanthology.org/W04-1013} {{ROUGE}: A package for automatic evaluation of summaries}.
\newblock In \emph{Text Summarization Branches Out}, pages 74--81, Barcelona, Spain. Association for Computational Linguistics.

\bibitem[{Liu et~al.(2021)Liu, Zheng, Demasi, Sabour, Li, Yu, Jiang, and Huang}]{esconv}
Siyang Liu, Chujie Zheng, Orianna Demasi, Sahand Sabour, Yu~Li, Zhou Yu, Yong Jiang, and Minlie Huang. 2021.
\newblock Towards emotional support dialog systems.
\newblock In \emph{Annual Meeting of the Association for Computational Linguistics}, pages 3469--3483.

\bibitem[{McHugh(2012)}]{kappa_range}
Mary~L McHugh. 2012.
\newblock Interrater reliability: the kappa statistic.
\newblock \emph{Biochemia medica}, 22(3):276--282.

\bibitem[{Mehrabi et~al.(2022)Mehrabi, Beirami, Morstatter, and Galstyan}]{MehrabiBMG22}
Ninareh Mehrabi, Ahmad Beirami, Fred Morstatter, and Aram Galstyan. 2022.
\newblock \href {https://doi.org/10.18653/V1/2022.NAACL-MAIN.204} {Robust conversational agents against imperceptible toxicity triggers}.
\newblock In \emph{Conference of the North American Chapter of the Association for Computational Linguistics: Human Language Technologies}, pages 2831--2847. Association for Computational Linguistics.

\bibitem[{{OpenAI}(2022)}]{ChatGPT2022}
{OpenAI}. 2022.
\newblock {Introducing ChatGPT}.
\newblock \url{https://openai.com/blog/chatgpt}.

\bibitem[{Ouyang et~al.(2022)Ouyang, Wu, Jiang, Almeida, Wainwright, Mishkin, Zhang, Agarwal, Slama, Ray et~al.}]{instruct-gpt}
Long Ouyang, Jeffrey Wu, Xu~Jiang, Diogo Almeida, Carroll Wainwright, Pamela Mishkin, Chong Zhang, Sandhini Agarwal, Katarina Slama, Alex Ray, et~al. 2022.
\newblock Training language models to follow instructions with human feedback.
\newblock volume~35, pages 27730--27744.

\bibitem[{Papineni et~al.(2002)Papineni, Roukos, Ward, and Zhu}]{bleu}
Kishore Papineni, Salim Roukos, Todd Ward, and Wei{-}Jing Zhu. 2002.
\newblock \href {https://doi.org/10.3115/1073083.1073135} {Bleu: a method for automatic evaluation of machine translation}.
\newblock In \emph{Annual Meeting of the Association for Computational Linguistics}, pages 311--318. {ACL}.

\bibitem[{Qiu et~al.(2023)Qiu, He, Zhang, Li, and Lan}]{smileconv}
Huachuan Qiu, Hongliang He, Shuai Zhang, Anqi Li, and Zhenzhong Lan. 2023.
\newblock Smile: Single-turn to multi-turn inclusive language expansion via chatgpt for mental health support.
\newblock \emph{arXiv preprint arXiv:2305.00450}.

\bibitem[{Radford et~al.(2018)Radford, Wu, Child, Luan, Amodei, and Sutskever}]{gpt2}
Alec Radford, Jeffrey Wu, Rewon Child, David Luan, Dario Amodei, and Ilya Sutskever. 2018.
\newblock \href {https://d4mucfpksywv.cloudfront.net/better-language-models/language-models.pdf} {Language models are unsupervised multitask learners}.

\bibitem[{Rashkin et~al.(2019)Rashkin, Smith, Li, and Boureau}]{empatheticdialogues}
Hannah Rashkin, Eric~Michael Smith, Margaret Li, and Y-Lan Boureau. 2019.
\newblock Towards empathetic open-domain conversation models: A new benchmark and dataset.
\newblock In \emph{Annual Meeting of the Association for Computational Linguistics}, pages 5370--5381.

\bibitem[{Roller et~al.(2021)Roller, Dinan, Goyal, Ju, Williamson, Liu, Xu, Ott, Smith, Boureau et~al.}]{blenderbot}
Stephen Roller, Emily Dinan, Naman Goyal, Da~Ju, Mary Williamson, Yinhan Liu, Jing Xu, Myle Ott, Eric~Michael Smith, Y-Lan Boureau, et~al. 2021.
\newblock Recipes for building an open-domain chatbot.
\newblock In \emph{Conference of the European Chapter of the Association for Computational Linguistics: Main Volume}, pages 300--325.

\bibitem[{Sharma et~al.(2020)Sharma, Miner, Atkins, and Althoff}]{SharmaMAA20}
Ashish Sharma, Adam~S. Miner, David~C. Atkins, and Tim Althoff. 2020.
\newblock \href {https://doi.org/10.18653/V1/2020.EMNLP-MAIN.425} {A computational approach to understanding empathy expressed in text-based mental health support}.
\newblock In \emph{Proceedings of the 2020 Conference on Empirical Methods in Natural Language Processing, {EMNLP} 2020, Online, November 16-20, 2020}, pages 5263--5276. Association for Computational Linguistics.

\bibitem[{Skilbeck and Payne(2003)}]{skilbeck2003emotional}
Julie Skilbeck and Sheila Payne. 2003.
\newblock Emotional support and the role of clinical nurse specialists in palliative care.
\newblock \emph{Journal of advanced nursing}, 43(5):521--530.

\bibitem[{Sun et~al.(2021)Sun, Lin, Zheng, Liu, and Huang}]{psyqa}
Hao Sun, Zhenru Lin, Chujie Zheng, Siyang Liu, and Minlie Huang. 2021.
\newblock Psyqa: A chinese dataset for generating long counseling text for mental health support.
\newblock In \emph{Findings of the Association for Computational Linguistics: ACL-IJCNLP}, pages 1489--1503.

\bibitem[{Touvron et~al.(2023)Touvron, Martin, Stone, Albert, Almahairi, Babaei, Bashlykov, Batra, Bhargava, Bhosale et~al.}]{llama2}
Hugo Touvron, Louis Martin, Kevin Stone, Peter Albert, Amjad Almahairi, Yasmine Babaei, Nikolay Bashlykov, Soumya Batra, Prajjwal Bhargava, Shruti Bhosale, et~al. 2023.
\newblock Llama 2: Open foundation and fine-tuned chat models.
\newblock \emph{arXiv preprint arXiv:2307.09288}.

\bibitem[{Tu et~al.(2022)Tu, Li, Cui, Wang, Wen, and Yan}]{tu2022misc}
Quan Tu, Yanran Li, Jianwei Cui, Bin Wang, Ji-Rong Wen, and Rui Yan. 2022.
\newblock Misc: A mixed strategy-aware model integrating comet for emotional support conversation.
\newblock In \emph{Annual Meeting of the Association for Computational Linguistics}, pages 308--319.

\bibitem[{Vincent J.~D'Andrea(1996)}]{j1996peer}
Peter~Salovey Vincent J.~D'Andrea. 1996.
\newblock Peer counseling-skills, ethics and perspectives.
\newblock \emph{Science and Behavior Books}, pages 29--36.

\bibitem[{von Werra et~al.(2020)von Werra, Belkada, Tunstall, Beeching, Thrush, and Lambert}]{vonwerra2022trl}
Leandro von Werra, Younes Belkada, Lewis Tunstall, Edward Beeching, Tristan Thrush, and Nathan Lambert. 2020.
\newblock Trl: Transformer reinforcement learning.
\newblock \url{https://github.com/huggingface/trl}.

\bibitem[{Wang et~al.(2023{\natexlab{a}})Wang, Wang, Mi, Deng, Wang, Liang, Xu, and Wong}]{wang2023cue}
Hongru Wang, Rui Wang, Fei Mi, Yang Deng, Zezhong Wang, Bin Liang, Ruifeng Xu, and Kam-Fai Wong. 2023{\natexlab{a}}.
\newblock Cue-cot: Chain-of-thought prompting for responding to in-depth dialogue questions with llms.
\newblock In \emph{Findings of the Association for Computational Linguistics: EMNLP}, pages 12047--12064.

\bibitem[{Wang et~al.(2023{\natexlab{b}})Wang, Li, and Zhao}]{wang2023self}
Jinyuan Wang, Junlong Li, and Hai Zhao. 2023{\natexlab{b}}.
\newblock Self-prompted chain-of-thought on large language models for open-domain multi-hop reasoning.
\newblock In \emph{Findings of the Association for Computational Linguistics: EMNLP}, pages 2717--2731.

\bibitem[{Wang et~al.(2023{\natexlab{c}})Wang, Li, Yang, Lin, and Wang}]{EEDG-PCI}
Lanrui Wang, Jiangnan Li, Chenxu Yang, Zheng Lin, and Weiping Wang. 2023{\natexlab{c}}.
\newblock Enhancing empathetic and emotion support dialogue generation with prophetic commonsense inference.
\newblock \emph{arXiv preprint arXiv:2311.15316}.

\bibitem[{Wang et~al.(2023{\natexlab{d}})Wang, Kordi, Mishra, Liu, Smith, Khashabi, and Hajishirzi}]{self-instruct}
Yizhong Wang, Yeganeh Kordi, Swaroop Mishra, Alisa Liu, Noah~A Smith, Daniel Khashabi, and Hannaneh Hajishirzi. 2023{\natexlab{d}}.
\newblock Self-instruct: Aligning language models with self-generated instructions.
\newblock In \emph{Annual Meeting of the Association for Computational Linguistics}, pages 13484--13508.

\bibitem[{Wei et~al.(2022)Wei, Wang, Schuurmans, Bosma, Xia, Chi, Le, Zhou et~al.}]{wei2022chain}
Jason Wei, Xuezhi Wang, Dale Schuurmans, Maarten Bosma, Fei Xia, Ed~Chi, Quoc~V Le, Denny Zhou, et~al. 2022.
\newblock Chain-of-thought prompting elicits reasoning in large language models.
\newblock \emph{Advances in Neural Information Processing Systems}, 35:24824--24837.

\bibitem[{Welivita and Pu(2023)}]{WelivitaP23}
Anuradha Welivita and Pearl Pu. 2023.
\newblock \href {https://doi.org/10.18653/V1/2023.FINDINGS-ACL.334} {Boosting distress support dialogue responses with motivational interviewing strategy}.
\newblock In \emph{Findings of the Association for Computational Linguistics: ACL}, pages 5411--5432. Association for Computational Linguistics.

\bibitem[{Yang et~al.(2023)Yang, Shi, Wan, Quan, Wang, Wu, and Wu}]{yang2023psycot}
Tao Yang, Tianyuan Shi, Fanqi Wan, Xiaojun Quan, Qifan Wang, Bingzhe Wu, and Jiaxiang Wu. 2023.
\newblock Psycot: Psychological questionnaire as powerful chain-of-thought for personality detection.
\newblock In \emph{Findings of the Association for Computational Linguistics: EMNLP}, pages 3305--3320.

\bibitem[{Zhang et~al.(2020)Zhang, Sun, Galley, Chen, Brockett, Gao, Gao, Liu, and Dolan}]{dialogpt}
Yizhe Zhang, Siqi Sun, Michel Galley, Yen-Chun Chen, Chris Brockett, Xiang Gao, Jianfeng Gao, Jingjing Liu, and William~B Dolan. 2020.
\newblock Dialogpt: Large-scale generative pre-training for conversational response generation.
\newblock In \emph{Annual Meeting of the Association for Computational Linguistics: System Demonstrations}, pages 270--278.

\bibitem[{Zhang et~al.(2022)Zhang, Zhang, Li, and Smola}]{zhang2022automatic}
Zhuosheng Zhang, Aston Zhang, Mu~Li, and Alex Smola. 2022.
\newblock Automatic chain of thought prompting in large language models.
\newblock In \emph{The Eleventh International Conference on Learning Representations}.

\bibitem[{Zheng et~al.(2023)Zheng, Sabour, Wen, Zhang, and Huang}]{augesc}
Chujie Zheng, Sahand Sabour, Jiaxin Wen, Zheng Zhang, and Minlie Huang. 2023.
\newblock Augesc: Dialogue augmentation with large language models for emotional support conversation.
\newblock In \emph{Findings of the Association for Computational Linguistics: ACL}, pages 1552--1568.

\end{thebibliography}

\appendix

\begin{table*}[ht]
\centering
\small
\scalebox{0.8}{
\begin{tabular}{c|l|p{14cm}}
\toprule
\multicolumn{1}{c}{} & \multicolumn{1}{c}{\textbf{Strategy}} & \multicolumn{1}{c}{\textbf{Definition \& Example}} \\
\midrule
1 & \textbf{Question$^\ast$} & Asking for information related to the problem to help the seeker articulate the issues they face and open-ended questions are best. 

\textit{E.g.: How do you define success in your life?}\\
\midrule
2 & \textbf{Restatement or Paraphrasing$^\ast$} & A simple, more concise rephrasing of the seeker’s statements that could help them see their situation more clearly. 

\textit{E.g.: It seems like you're experiencing a crisis of faith and that it's causing you a lot of inner turmoil and confusion.}\\
\midrule
3 & \textbf{Reflection of Feelings$^\ast$} & Articulate and describe the seeker’s feelings. 

\textit{E.g.: It seems like you're feeling quite anxious about this decision.}\\
\midrule
4 & \textbf{Self-disclosure$^\ast$} & Divulge similar experiences the supporter has had or emotions the supporter shares with the seeker to express empathy. 

\textit{E.g.: I've felt similar when I was in a tough spot at work; it's really challenging.}\\
\midrule
5 & \textbf{Affirmation and Reassurance$^\ast$} & Affirm the seeker’s strengths, motivation, and capabilities and provide reassurance and encouragement. 

\textit{E.g.: You've been really resourceful in handling challenges before; I'm confident you can get through this.}\\
\midrule
6 & \textbf{Providing Suggestions$^\ast$} & Provide suggestions about how to change, but be careful not to overstep and tell them what to do. 

\textit{E.g.: Perhaps trying a new approach to this problem might yield different results. Have you considered brainstorming with a team?}\\
\midrule
7 & \textbf{Information$^\ast$} & Provide useful information to the seeker, e.g., data, facts, opinions, resources, or by answering questions. 

\textit{E.g.: There are several techniques for stress management, like mindfulness and exercise, that may be helpful.}\\
\midrule
8 & \textbf{Summarize} & Brief recaps that highlight key themes from both the seeker's and the supporter's interactions, and make connections within the seeker's story. 

\textit{E.g.: So it sounds like you're struggling to manage your debts and you feel overwhelmed by the situation. We talked about some practical steps you can take, such as reaching out to your lenders and setting up a budget. You also mentioned feeling alone and helpless, but I want to remind you that you're not alone and there are resources and support available to you.}\\
\midrule
9 & \textbf{Imagery} & Encourage the seeker to visualize different situations or outcomes to gain insights or relieve stress. 

\textit{E.g.: Imagine a scenario where you've successfully overcome your current challenges. How does that look and feel?}\\
\midrule
10 & \textbf{Specify} & Invite the seeker to provide more detail on the general statements they have previously made, for example, by using concrete instances. 

\textit{E.g.: You mentioned feeling overwhelmed. Can you give a specific example of what triggers this feeling?}\\
\midrule
11 & \textbf{Take Responsibility} & Encourage the seeker to take suitable responsibility for their actions. 

\textit{E.g.: Ultimately, the decision is yours to make. You have to take responsibility for the choice you make and how it impacts your future.}\\
\midrule
12 & \textbf{Homework Assignment} & Given tasks or activities to help the seeker practice what they learned in the conversation. 

\textit{E.g.: For the next week, try to jot down your thoughts in a journal whenever you feel stressed. It might help you understand your emotions better.}\\
\midrule
13 & \textbf{Immediacy} & Involve the supporter in sharing their immediate feelings about the seeker or their relationship. 

\textit{E.g.: I want you to know that I'm here for you if you need anything else.}\\
\midrule
14 & \textbf{Others} & Exchange pleasantries and use other support strategies that do not fall into the above categories. \\
\bottomrule
\end{tabular}
}
\caption{Definitions of enriched strategies. $\ast$ represents the strategy also used in ESConv.}
\label{tab:strategy_defi}
\end{table*}

\section{Details of ESD Dataset Construction}
\label{Appdix:ESD}

\subsection{Prompt for Situation Generation}
\label{Appdix:situation-prompt}
\begin{figure}[h]
\centering
\includegraphics[scale=0.73]{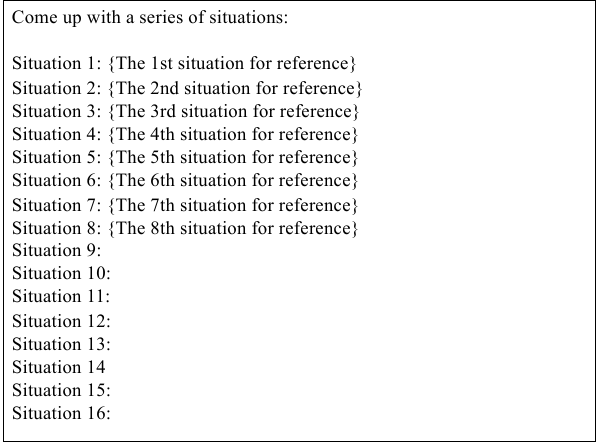}
\caption{The prompt used for generating new situations.}
\label{fig:situation_prompt}
\end{figure}

Inspired by~\citet{instruct-gpt}, we use the prompt in Figure~\ref{fig:situation_prompt} to generate new situations.

\begin{table}[tbp]
  \centering
  \small
  \vspace{-10pt}
  \scalebox{0.91}{
  {
    \begin{tabular}{p{25em}}
    \toprule
    \multicolumn{1}{c}{\textbf{ESConv}} \\
    \midrule  lol (5.1), covid (5.1), bye (4.9), vaccine (4.5), wow (4.3), mad (4.2), guy (4.2), virus (4.2), holiday (4.2), yea (4.2), press (4.1), girl (4.0), sucks (4.0), horrible (4.0), pray (3.9), fine (3.8), super (3.8), bet (3.8), alot (3.8), counselling (3.8), grade (3.7), lockdown (3.7), boyfriend (3.7), zoom (3.7), survey (3.7), ya (3.6), awhile (3.6), gonna (3.6), closed (3.6), 19 (3.5)   \\
    \midrule[0.3mm]
    \multicolumn{1}{c}{\textbf{ESD}} \\
    \midrule    visualizing (5.7), discrimination (5.4), resilience (5.2), visualize (5.1), visualization (5.0), caregiving (4.9), exploring (4.7), summary (4.6), specializes (4.5), infertility (4.5), incorporating (4.4), vision (4.4), bravery (4.3), unfulfilled (4.3), caregivers (4.3), gender (4.2), navigating (4.2), sexuality (4.2), accommodations (4.2), drowning (4.1), prioritize (4.0), empowered (4.0), behavioral (4.0), effectively (4.0), substance (4.0), cognitive (4.0), culture (3.9), covers (3.9), align (3.9), setbacks (3.9) \\
    \bottomrule
    \end{tabular}%
    }
    }
  \caption{
  Top 30 salient topic features. The values in parentheses are the z-scored log odds ratios for the corresponding word. 
  }
  \label{tab:log_odds_ratio_zscore}
\end{table}%

\subsection{Definitions and Examples of Enriched Strategies}
\label{Appdix:strategy}
Through our enriching process, we double the size of strategies based on ESConv. Table~\ref{tab:strategy_defi}  presents the entire list of strategies that are employed in our emotional support dialogue generation, including both definition and example for each strategy. 

\subsection{Details of Dialogue Diversity Analysis} 
\label{Appdix:ESD-diversity}
We calculate the log odds ratio values for the words of ESD relative to ESConv, which is formulated as follows:
\begin{equation}
\small
Log Odds Ratio(i) = log ((p_{i,1}/(1-p_{i,1}))/(p_{i,2}/(1-p_{i,2}))
\end{equation}
where \textit{$p_{i,1}$} and \textit{$p_{i,2}$} represent the probabilities of the word \textit{i} in two compared datasets.
Based on the log odds ratio values, we calculate the z-score as follows:
\begin{equation}
\small
z\text{-}score(i) = (Log Odds Ratio(i)-\mu)/\sigma
\end{equation}
where {$\mu$} is the mean of all words' log odds ratio, and {$\sigma$} is the standard deviation.

We rank all words based on their z-scores. The top 25 significant words and their z-scores for each dataset are shown in Table~\ref{tab:log_odds_ratio_zscore}. In the comparison between two datasets, the word with higher z-scores can represent a greater distinction within its dataset compared to the other dataset.

In the ESConv dataset, words such as `wow', `mad', `sucks', and `horrible' reflect emotional states and reactions. Additionally, word like `counselling', `grade', `zoom' and `survey' point to various types events in school. Words like `covid', `vaccine', `virus', and `holiday' suggest discussions about time and significant events. Expressions like `lol', `bye', and `ya' indicate informal and colloquial conversations.

In contrast, our dataset contains words like `resilience', `visualizing', `caregiving', and `vision', which relate to coping strategies and empowerment. Words like `gender', `sexuality', and `accommodations' indicate a theme related to identity and inclusivity. The presence of words like `substance', `cognitive', and `culture' suggest that there are discussions focusing on psychological and cultural aspects in ESD dataset.

The analysis of these significant words shows that ESConv dataset are more general, focusing on everyday life, emotions, and informal interactions. In contrast, our dataset is more specific and related to more professional guidance, diving into subjects related to mental health and personal challenges deeper.

\subsection{Prompts for Strategy Impact Assessment}
\label{Appdix:ESD-StratImpt}
\begin{figure}[t]
\centering
\includegraphics[scale=0.65]{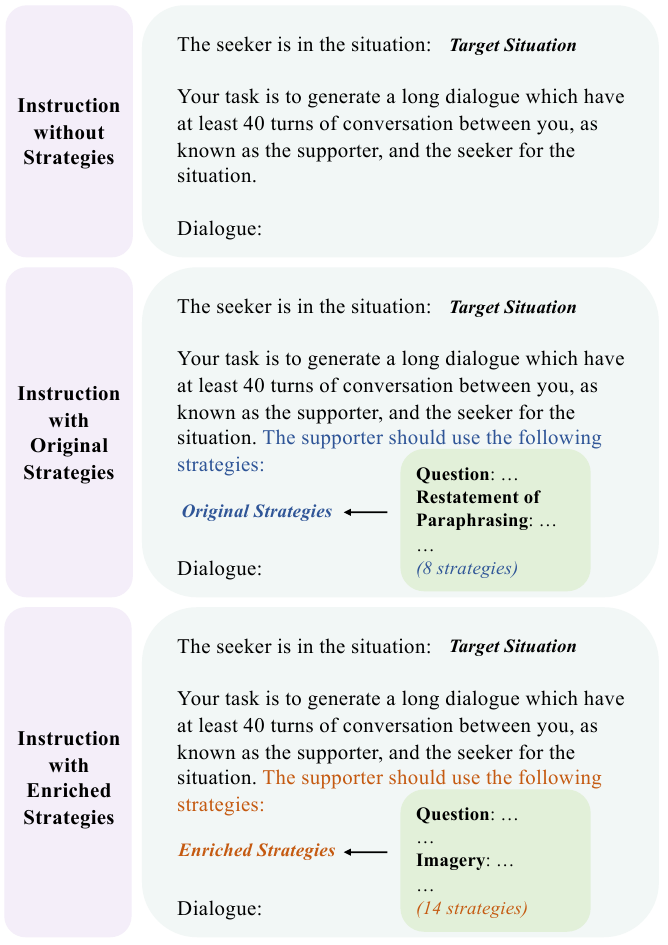}
\caption{The instruction part of prompts used for the strategy impact assessment. The differences between the instruction parts are highlighted.}
\label{fig:prompt_strategy_impact_assessment}
\end{figure}

To assess the impact of incorporating strategies into prompts on the quality of generated dialogues, we design three different prompts: (a) without adding strategies, (b) incorporating strategies solely from ESConv, and (c) adding our enriched strategies. Only the instruction parts of these prompts are different, which are outlined in Figure~\ref{fig:prompt_strategy_impact_assessment} with the differences highlighted. 

\section{Chain Creation Prompt of ESD-CoT Dataset}
\label{Appdix:ESD-CoT}
\begin{figure}[t]
\centering
\includegraphics[scale=0.65]{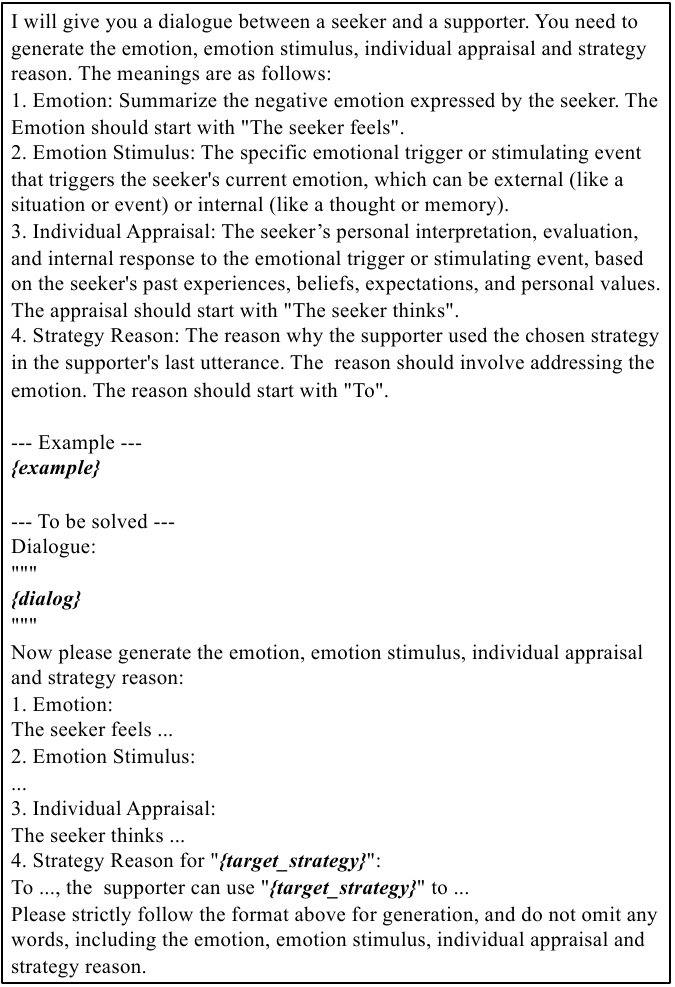}
\caption{The prompt template used for chain creation of the ESC-CoT dataset.}
\label{fig:cot_prompt}
\end{figure}

The prompt template used for generating chains is shown in Figure~\ref{fig:cot_prompt}. 
We first introduce the task and give the definition of each element.
Then, we provide an \textbf{\textit{example}}, which will be replaced with the chain example in the example pool.
Furthermore, we provide the \textbf{\textit{dialogue}}, which needs to be supplemented with the chain.
Finally, we standardize the format of the output. Note the \textbf{\textit{target strategy}} will be replaced with the strategy used in the formerly generated dialogue. 

\section{Details of Human Evaluation}
\label{Appdix:humaneval}
As for the response evaluation, to fairly evaluate different responses, we randomly shuffle the responses when evaluated by annotators. We pay for these 3 annotators, and all the results are proof-checked by an expert.

As for the strategy consistency evaluation, we randomly choose 100 chains for the settings containing \(SR\) and recruit 2 professional annotators to evaluate the consistency and another professional annotator to deal with the situation where two annotators’s assessments are inconsistent.

\section{Implementation Details}
\label{Appdix:Details}

\subsection{Supervised Fine-Tuning} 
\label{Appdix:Details-SFT}

\paragraph{BlenderBot}
is an open-domain conversational agent equipped with a range of communication skills, including empathetic response capabilities. For our experiments, we use the small\footnote{https://huggingface.co/facebook/BlenderBotbot\_small-90M} version of BlenderBot.
We train the BlenderBot model based on the code of ESConv\footnote{https://github.com/thu-coai/Emotional-Support-Conversation}~\cite{esconv}.
We train the model using one A6000 GPU, with a batch size of 64, a learning rate of 3e-5, and a max sequence length of 500.

\paragraph{DialoGPT}
is a model built upon the foundation of the dialogue generative pre-trained transformer, specially, GPT-2~\cite{gpt2}. For our experiments, we use the small\footnote{https://huggingface.co/microsoft/DialoGPT-small} version of DialoGPT.
We train the DialoGPT model based on the code of ESConv~\cite{esconv}.
We train the model using one A6000 GPU, with a batch size of 32, a learning rate of 5e-5, and a max sequence length of 500.

\paragraph{\textsc{Llama2-Chat}}
\label{llama2-sft}
is a pre-trained model optimized for dialogue use cases, which has been specifically designed to adapt to a wide range of conversational scenarios. For our experiments, we use the 7B version of \textsc{Llama2-Chat} on Hugging Face.
The training of \textsc{Llama2-Chat} model is based on the SFT trainer of Transformer Reinforcement Learning\footnote{https://github.com/huggingface/trl}~\cite{vonwerra2022trl}.
We train the model using 4*A6000 GPUs, with batch size of 8 per GPU, learning rate of 5e-5, and max sequence length of 2048.

\subsection{Ablation Study}
\label{Appdix:Details-ablation}
We conduct the ablation study based on \textsc{Llama2-Chat}. 
The model size and parameters are the same as those in Appendix~\ref{llama2-sft}.
The only difference between different settings is the composition of the data.
For example, in the setting of $\{EM, SR, RE\}$, the instruction is "\textit{Generate the response as the supporter using the pipeline of Emotion, Strategy Reason, Response.}" and the corresponding ground truth consists of the manually checked \textit{Emotion}, \textit{Strategy Reason}, and \textit{Response}.
For each setting, we train the model for 10 epochs using 4*A6000 GPUs for approximately 3 hours, with batch size of 8 per GPU, learning rate of 5e-5, and max sequence length of 2048 on the training dataset, and select the best-performing checkpoint based on the validation dataset to obtain its performance metrics on the test dataset.

\section{Examples of ESD Dataset}
\label{Appdix:ESDExamples}
\begin{figure*}[t]
\centering
\includegraphics[scale=0.57]{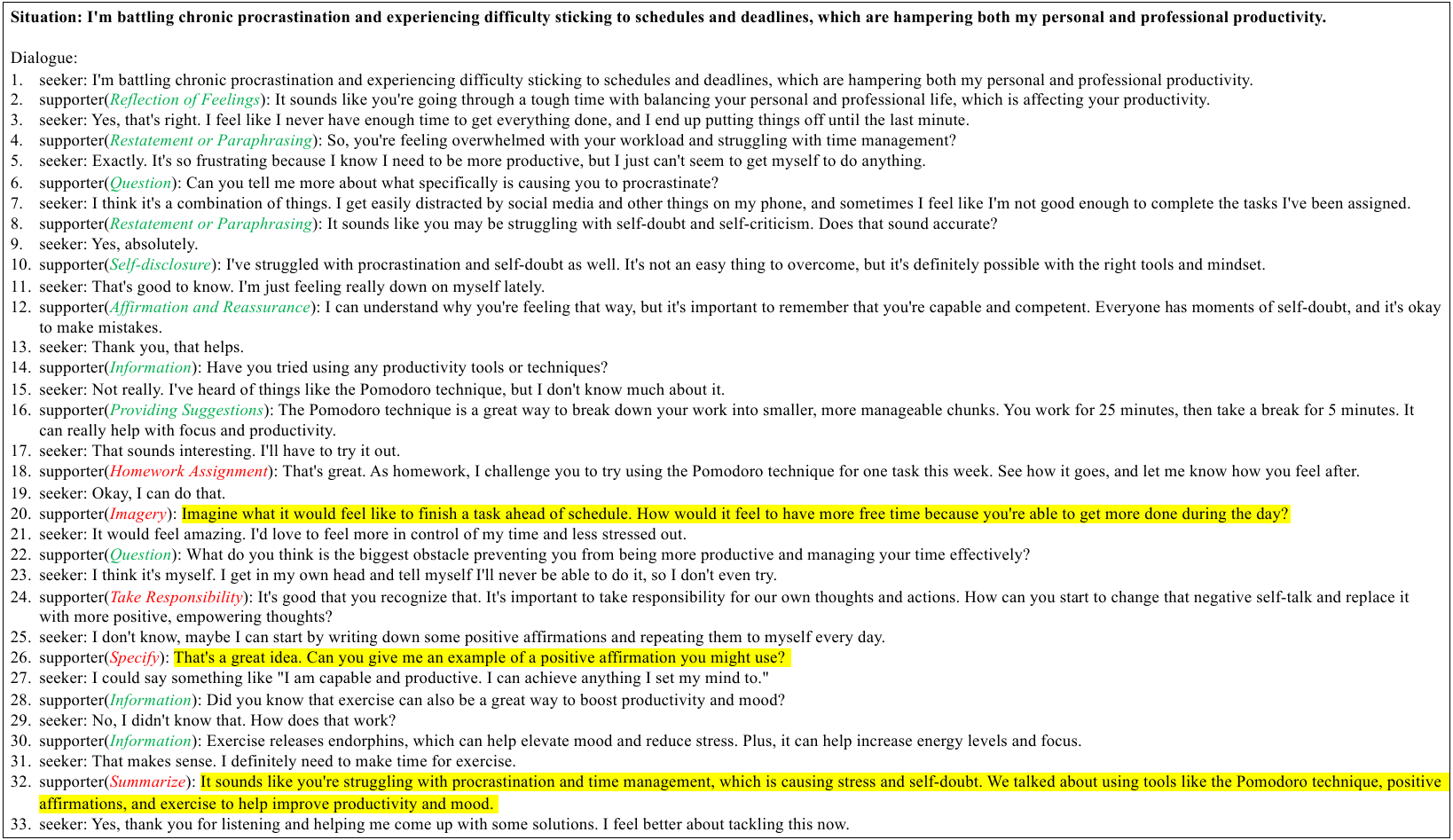}
\caption{An example of the ESD dataset. The sentences corresponding to the \textcolor{myred}{enriched strategies} are \textbf{notably distinct} from the \textcolor{mygreen}{original strategies}. 
For instance, the sentence in line 20 makes the seeker see the hope and create new experiences by \textit{Imagery} strategy. 
Then, the sentence in line 26 makes the seeker's methods more clearly and easier to implement by \textit{Specify} strategy, 
Furthermore, the sentence in line 32 provides a comprehensive overview of the current challenges and solutions by \textit{Summarize} strategy. 
These aspects were relatively uncommon in previous emotional support dialogue datasets.}
\label{fig:case1}
\end{figure*}

\begin{figure*}[t]
\centering
\includegraphics[scale=0.57]{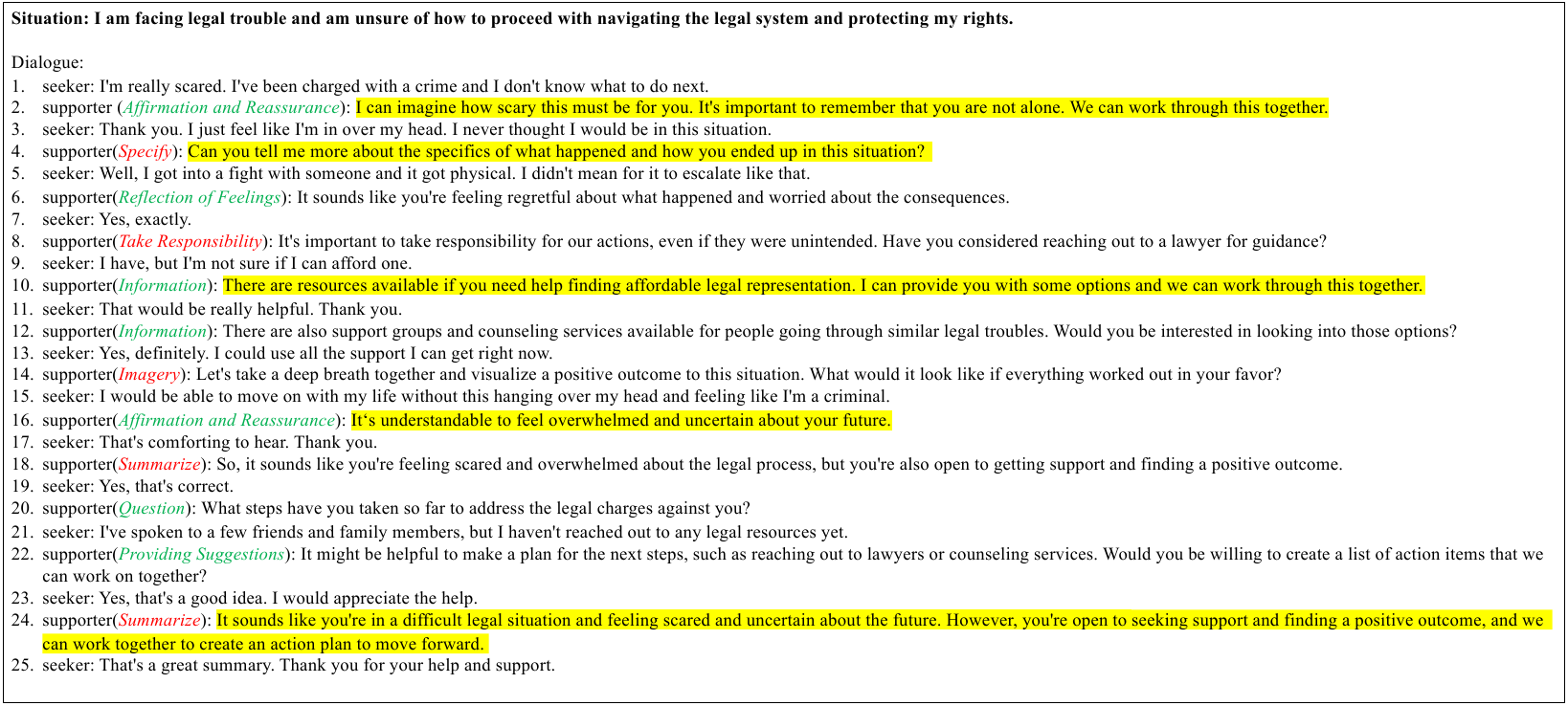}
\caption{Another example of the ESD dataset. The sentences corresponding to the \textcolor{mygreen}{original strategies} and \textcolor{myred}{enriched strategies} \textbf{both contribute} to the progression of the dialogue. 
For instance, the sentence in line 2 timely responds to and reassures the scare exposed by the seeker is seen by the supporter in line 1. 
The sentence in line 4 asks the seeker to reveal crucial specific information so that the current situation will be more clear, and gives the seeker an opportunity to express. 
The sentence in line 10 responds to the difficulty of the seeker in line 9 and clarifies that the supporter can offer useful resources, which gives the direction for the subsequent content. 
The sentence in line 16 affirms and stabilizes the seeker's emotions. 
Finally, the sentence in line 24 summarizes the entire dialogue and proposes further arrangements, enabling the seeker not to fall into the current situation.
}
\label{fig:case2}
\end{figure*}

We present examples of the ESD dataset in Figure~\ref{fig:case1} and Figure~\ref{fig:case2}, including situations and the generated dialogues. The strategies originally present in ESConv are labeled in green, while the strategies we enriched are marked in red.

\section{Examples of ESD-CoT Dataset}
\label{Appdix:ESDCoTExamples}
\begin{figure*}[t]
\centering
\includegraphics[scale=0.8]{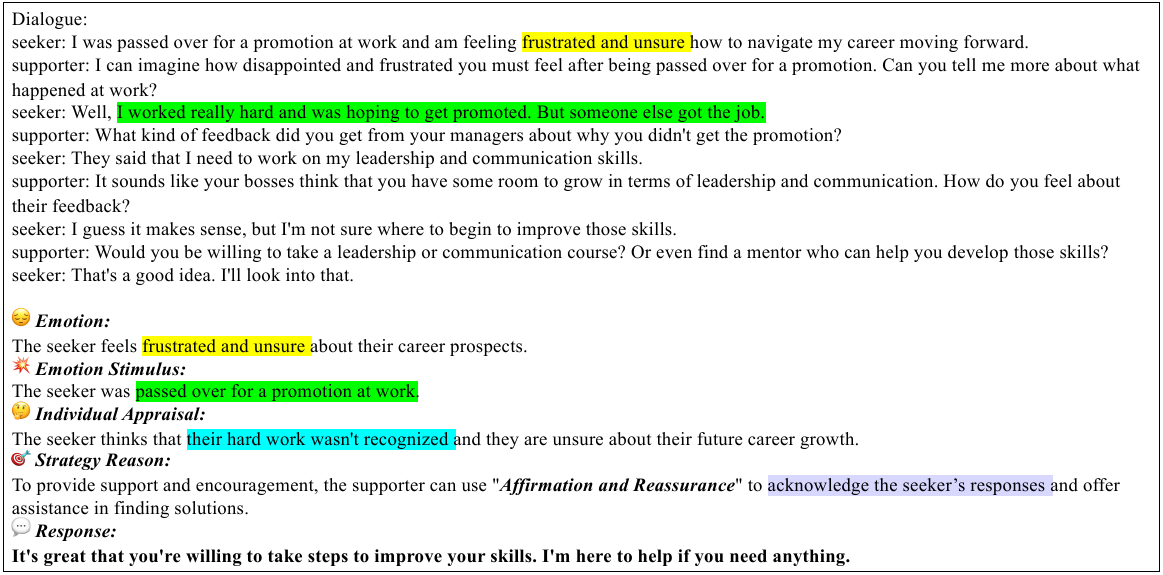}
\caption{An example of the ESD-CoT dataset. The emotion and stimulus correspond to what the seeker said. The individual appraisals of ``they don't know'' and ``feel the need'' reflect the seeker's appraisal of the situation. The description of the strategy reason is reasonable, and the strategy and the response are also consistent.
}
\label{fig:cot_case1}
\end{figure*}

\begin{figure*}[ht]
\centering
\includegraphics[scale=0.8]{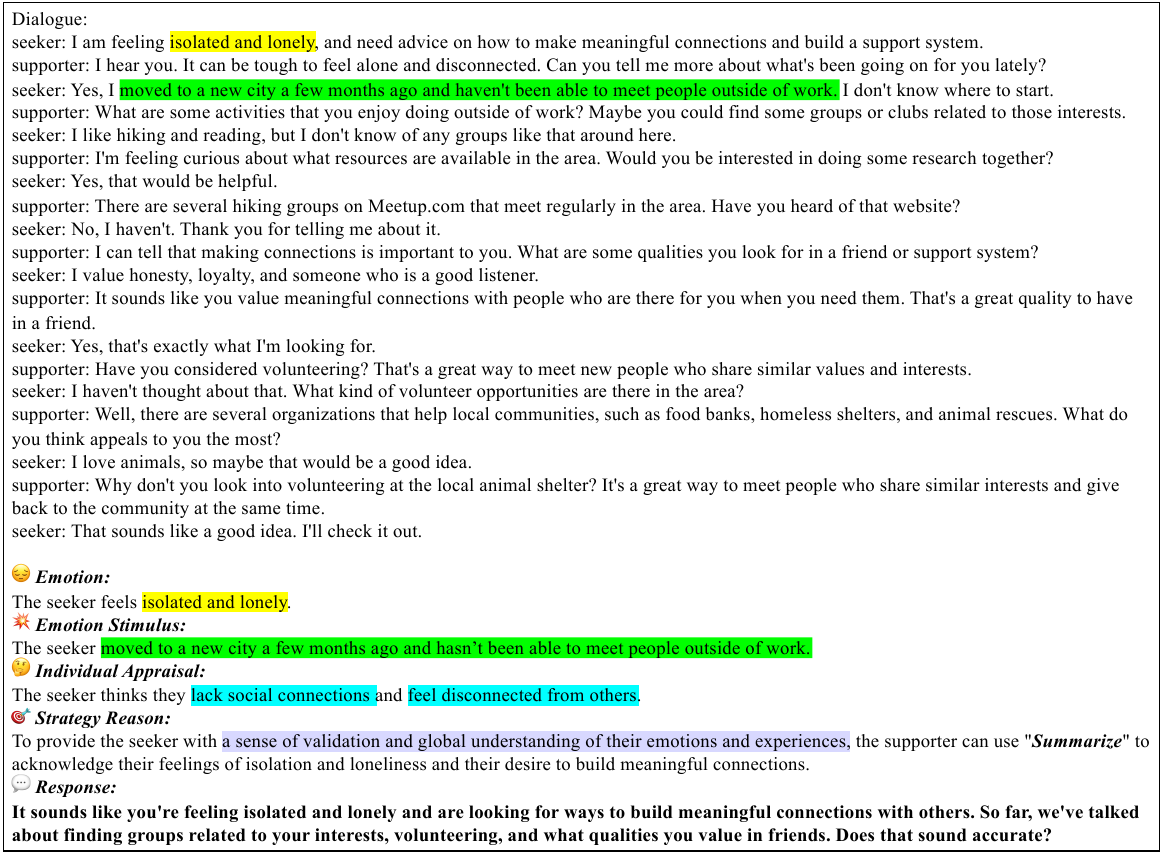}
\caption{Another example of the ESD-CoT dataset, which uses the strategy of ``\textbf{\textit{Summarize}}''. 
}
\label{fig:cot_case2}
\end{figure*}

We present examples of the ESD-CoT dataset in Figure~\ref{fig:cot_case1} and Figure~\ref{fig:cot_case2}.
The upper part is the context of the dialogue, and the lower part is the chain to generate the supporter's response.
The chain and the corresponding part of the dialogue are highlighted in the same color.

\section{Case Study}
\label{Appdix:case_study}
In Table~\ref{tab:case_study}, we demonstrate a case of the ablation study. In this case, the seeker feels trapped and isolated in a toxic relationship and unsure about what the right thing to do is. The model's responses from different incomplete chain settings are focused only on one aspect. In the full chain setting, the model's response considers empathy and emphasizes the importance of personal responsibility. It encourages the seeker to make decisions that align with their values and goals and to clarify the next steps. Besides, the reason for the generated response can be clearly seen in the setting of the entire chain.

\begin{table*}[ht]
\small
\centering
\begin{tabular}{c|l}
\toprule
\textbf{Situation} & I am in a toxic or abusive relationship and am unsure of how to leave safely or how to seek help \\
\midrule
\multirow{13}{*}{\textbf{Context}} & \textbf{Supporter}: Hi there, how are you feeling today? \\
 & \textbf{Seeker}:  Not so great. I'm feeling really trapped in my relationship. \\
 & \textbf{Supporter}: Can you tell me more about that? What's happening? \\
 & \textbf{Seeker}: My partner is really controlling, and I feel like I can't do anything without their approval. I'm\\
 & starting to feel really isolated from my family and friends. \\
 & \textbf{Supporter}: It sounds like you're feeling trapped and controlled in your relationship, and that's leading\\
 & to feelings of isolation. \\
 & \textbf{seeker}: Yes, exactly. I don't know what to do.\\
 & \textbf{Supporter}: If you're feeling like you're in a dangerous situation, you can call the National Domestic\\
 & Violence Hotline at 1-800-799-7233 for immediate support and next steps.\\
 & \textbf{seeker}: Thank you, I'll keep that in mind.\\
 & \textbf{Supporter}: Based on what you're saying, it sounds like you're feeling trapped and isolated because of\\
 & your partner's controlling behavior. You're not sure what to do next. \\
 & \textbf{seeker}: Yes, that's right. It's hard to know what the right thing to do is.\\
\midrule
\textbf{RE} & Have you talked to your partner about how you're feeling? \\
\textbf{SR\_RE} & It sounds like you're feeling really stuck in this situation, and it's hard to see a way out. \\
\textbf{EM\_SR\_RE} & Imagine a future where you're no longer trapped and isolated. What does that look like to you?\\
\midrule
\multirow{12}{*}{\textbf{ESCoT}} 
& \textcolor[rgb]{0, 0.7, 0}{\textbf{Emotion}}: The seeker feels \textbf{trapped and isolated}. \\
& \textcolor[rgb]{0, 0.7, 0}{\textbf{Emotion Stimulus}}: The seeker's \textbf{partner is controlling} and the seeker feels like they \textbf{can't do any}\\
& \textbf{-thing without their approval}. \\
& \textcolor[rgb]{0, 0.7, 0}{\textbf{Individual Appraisal}}: The seeker thinks that \textbf{their partner's controlling behavior is making them}\\
& \textbf{feel trapped and isolated}, and they are \textbf{unsure of what to do next}. \\
& \textcolor[rgb]{0, 0.7, 0}{\textbf{Strategy Reason}}: To provide the seeker with a sense of empowerment and support, the supporter can\\
& use \textbf{"Take Responsibility"} to encourage the seeker to take control of their own life and make decisions\\
& that align with their values and goals. This strategy aims to address the seeker's feelings of being trapped\\
& and isolated by emphasizing the importance of taking responsibility for their own actions and well-being.\\
& \textcolor[rgb]{0, 0.7, 0}{\textbf{Response}}: It sounds like you're feeling really stuck in this situation, but I want you to know that you\\
& have the power to take control of your own life. You can make decisions that align with your values and\\
& goals, and that will help you feel more empowered and in control. \\
\midrule
\multirow{2}{*}{\textbf{Ground-Truth}} & It's important to remember that you're not alone, and that you deserve to feel supported and safe\\
& in your relationships. \\
\bottomrule
\end{tabular}

\caption{A case of the generated supportive responses by the ablation study. We show the model's responses in different experimental settings, "RE", "SR\_RE", and "EM\_SR\_RE". In the case of "ESCoT", the entire chain setting, we display the output of all elements.}
\label{tab:case_study}
\end{table*}

\end{document}